\documentclass[twoside]{article}

\usepackage[accepted]{aistats2025}
\usepackage[round]{natbib}

\usepackage[utf8]{inputenc} 
\usepackage[T1]{fontenc}    
\usepackage{hyperref}       
\usepackage{url}            
\usepackage{booktabs}       
\usepackage{amsfonts}       
\usepackage{nicefrac}       
\usepackage{microtype}      
\usepackage[dvipsnames]{xcolor}         
\hypersetup{
    colorlinks=true,
    citecolor=Blue,
    linkcolor=red,
    urlcolor=orange,
}
\usepackage{amsmath}
\usepackage{amsthm}
\usepackage{amssymb}
\usepackage{mathtools}
\usepackage{bm}
\usepackage{algorithmic}
\usepackage{algorithm}
\usepackage{wrapfig}
\usepackage{lipsum}
\usepackage{bbm}
\usepackage{setspace}
\newtheorem{lemma}{Lemma}
\newcommand{\rom}[1]{\romannumeral #1\relax}
\usepackage{mathbbol}
\usepackage[T1]{fontenc}
\usepackage{lmodern}

\DeclareSymbolFontAlphabet{\mathbb}{AMSb}
\DeclareSymbolFontAlphabet{\mathbbl}{bbold}

\usepackage{setspace}

\begin{document}

%

%

\twocolumn[

\aistatstitle{Hyperboloid GPLVM for Discovering Continuous Hierarchies via Nonparametric Estimation}

\aistatsauthor{Koshi Watanabe \And Keisuke Maeda \And Takahiro Ogawa \And Miki Haseyama}
\aistatsaddress{Hokkaido University \And Hokkaido University \And Hokkaido University \And Hokkaido University}
]

\begin{abstract}
Dimensionality reduction (DR) offers interpretable representations of complex high-dimensional data, and recent DR methods have leveraged hyperbolic geometry to obtain faithful low-dimensional embeddings of high-dimensional hierarchical relationships. However, existing methods are dependent on neighbor embedding, which frequently ruins the continuous nature of the hierarchical structures. This paper proposes hyperboloid Gaussian process latent variable models (hGP-LVMs) to embed high-dimensional hierarchical data while preserving the implicit continuity via nonparametric estimation. We adopt generative modeling using the GP, which provides effective hierarchical embedding and executes ill-posed hyperparameter tuning. This paper presents three variants of the proposed models that employ original point, sparse point, and Bayesian estimations, and we establish their learning algorithms by incorporating the Riemannian optimization and active approximation scheme of the GP-LVM. In addition, we employ the reparameterization trick for scalable learning of the latent variables in the Bayesian estimation method. The proposed hGP-LVMs were applied to several datasets, and the results demonstrate their ability to represent high-dimensional hierarchies in low-dimensional spaces.
\end{abstract}

\section{Introduction}
With the emergence of large high-dimensional datasets, unsupervised dimensionality reduction (DR) techniques have gained increasing attention in discovering a faithful low-dimensional representation while preserving essential characteristics of the data. Recent studies have shifted focus from conventional toy datasets to more complicated datasets, e.g., neural activities~\citep{jensen2020manifold} or single-cell ribonucleic acid sequencing (scRNA-seq)~\citep{becht2019dimensionality}, frequently leveraging Riemannian geometry to achieve effective data embedding. Nonlinear hierarchical relationships are frequently encountered in high-dimensional data, where \textit{hyperbolic embedding}~\citep{nickel2017poincare, nickel2018learning} has proven particularly effective. Utilizing curved hyperbolic geometry, hyperbolic embedding enables faithful DR of hierarchical datasets while requiring far fewer dimensions than their Euclidean counterparts~\citep{sala2018representation}. However, to the best of our knowledge, visualization-aided hyperbolic embedding~\citep{klimovskaia2020poincare, jaquier2022bringing} remains relatively underexplored despite its applicability in visualizing high-dimensional hierarchical data. Thus, this paper focuses on advancing hyperbolic embedding to realize improved visualization of hierarchical data structures.
\par
\begin{figure*}
    \centering
    \includegraphics[width=160mm]{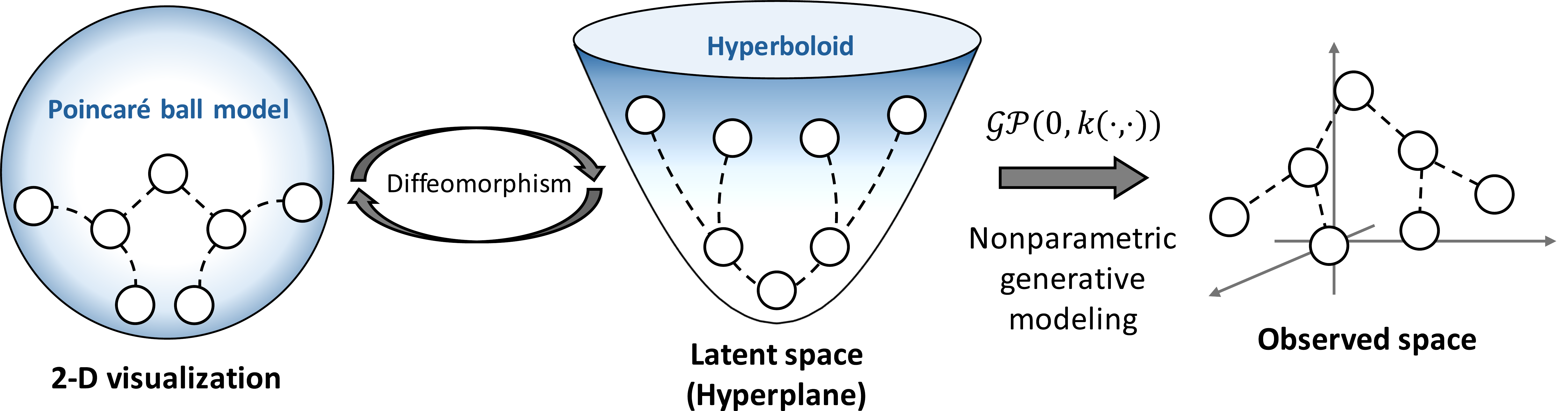}
    \caption{\textbf{Illustration of hyperboloid Gaussian process latent variable models (hGP-LVMs)}. We learn the latent variables on the Lorentz model and visualize them on the Poincar{\'e} ball model.}
    \label{fig:abstract}
\end{figure*}
First, we categorize previous DR methods into four classes based on two key axes to establish effective DR of hierarchical data, i.e., \textit{parametric} vs. \textit{nonparametric} and \textit{data-embedding} vs. \textit{neighbor-embedding}. Specifically, we classify DR methods as \textit{parametric} or \textit{nonparametric} based on whether they involve a parametric projection between the observed and latent spaces. Similarly, we distinguish between \textit{data-embedding} and \textit{neighbor-embedding} methods depending on whether they utilize the raw data or rely on neighbor relations (e.g., a $k$-nearest neighbor graph). For example, principal component analysis (PCA)~\citep{hotelling1933analysis} and variational autoencoders (VAE)~\citep{kingma2013auto, higgins2017betavae} are parametric data-embedding methods, and t-stochastic neighbor embedding (t-SNE)~\citep{van2008visualizing} and uniform manifold approximation and projection (UMAP)~\citep{mcinnes2018umap} are \textit{nonparametric neighbor-embedding} methods. 
\par
Conventionally, visualization-aided DR is based on nonparametric neighbor-embedding methods because they are not dependent on ill-posed parameter tuning and better preserve local structures. However, for hierarchical data embedding, the neighbor embedding methods frequently create discrete clusters~\citep{amid2019trimap, wang2021understanding}, potentially overlooking the global \textit{implicit continuity} of the hierarchical structure. This inherent limitation of neighbor-embedding methods has motivated the development of \textit{nonparametric data-embedding} methods, which are better suited for DR of hierarchical data.
\par
Gaussian process (GP) latent variable models (GP-LVMs)~\citep{lawrence2005probabilistic,titsias2010bayesian} are representative examples of nonparametric data-embedding methods. GP-LVMs assume the GP decoder of the observed variables, which is a widely used nonparametric model to estimate unknown functions~\citep{rasmussen2006gaussian}. Generative modeling between the observed and latent spaces frequently enhances the continuity of the latent variables, thereby making them suitable for preserving the continual relationships in hierarchical data. However, recent studies have explored incorporating simple Riemannian geometries in the GP~\citep{mallasto2018wrapped, borovitskiy2020matern} or developing supervised kernel methods in hyperbolic spaces~\citep{fang2021kernel, fan2023horospherical}, and only a few methods realize the hyperbolic extension of GP-LVMs~\citep{jaquier2022bringing}.
\par
This paper proposes \textit{hyperboloid GP-LVMs} (\textit{hGP-LVMs}) to visualize the hierarchical relationships behind high-dimensional data via nonparametric estimation (Figure~\ref{fig:abstract}). We learn the latent variables on the Lorentz model and visualize them on the Poincar\'{e} ball model by applying the diffeomorphism among them, and we learn the hyperbolic latent variables by developing dedicated algorithms that incorporate the previous Riemannian optimization and sparse GP methods. In addition, we formulate three variants of hGP-LVMs, i.e., extension of the original point estimation~\citep{lawrence2005probabilistic}, sparse point estimation~\citep{lawrence2007learning, titsias2009variational}, and Bayesian estimation methods~\citep{titsias2010bayesian, lalchand2022generalised} to address classical computational issues associated with the GP and realize scalable Bayesian learning of latent variables. The primary contribution of this paper is the development of the hGP-LVMs to visualize hierarchical data effectively via nonparametric estimation using generative modeling with the GP decoder. We demonstrate that the GP-based modeling and Bayesian estimation of the latent variables benefit visualization-aided DR on both synthetic and real-world datasets.

\section{Background}
\label{section:Backgrounds}
Before introducing the proposed hGP-LVMs, we first discuss previous GP-LVMs and the basic concepts of the hyperbolic space.
\subsection{Gaussian Process Latent Variable Models}
Here, let $\bm{\mathrm{y}}_{i}\in\mathbb{R}^{D}\,(i=1,2,\ldots,N)$ be the $D$-dimensional observed variables and let $\bm{\mathrm{x}}_{i}\in\mathcal{M}^{Q}$ be the latent variables on a $Q$-dimensional smooth manifold $\mathcal{M}^{Q}$, i.e., the target of DR. We denote $\bm{\mathrm{Y}}=[\bm{\mathrm{y}}_{1},\bm{\mathrm{y}}_{2},\ldots,\bm{\mathrm{y}}_{N}]^{\top}\in\mathbb{R
}^{N \times D}$ and $\bm{\mathrm{X}}=[\bm{\mathrm{x}}_{1},\bm{\mathrm{x}}_{2},\ldots,\bm{\mathrm{x}}_{N}]^{\top}\in\mathbb{R
}^{N \times Q}$, and let $E$ be an Euclidean manifold. The original GP-LVM is given as follows: 
\begin{align}
    \bm{\mathrm{y}}_{:,d}&=\bm{\mathrm{f}}_{d}(\bm{\mathrm{X}})+\bm{\epsilon}\label{eq:GP_regression},\\
    \;\bm{\epsilon}&\sim\mathcal{N}(\bm{0},\beta^{-1}\bm{\mathrm{I}}_{n})\label{eq:GP_noise},\\
    \bm{\mathrm{f}}_{d}&\sim\mathcal{GP}(\bm{0},k(\cdot,\, \cdot)), \label{eq:gaussian_process}
\end{align}
where $\bm{\mathrm{f}}_{d}\in\mathbb{R}^{N}\,(d=1,2,\ldots,D)$ is a GP prior with a kernel function $k(\cdot,\cdot)$, and $\bm{\epsilon}\in\mathbb{R}^{N}$ is Gaussian noise with a precision $\beta$. The GP-LVM estimates the latent variables by maximizing the log-likelihood $\mathcal{F}=\sum_{d=1}^{D}\log p(\bm{\mathrm{y}}_{:,d}|\bm{\mathrm{X}})$ as follows:
\begin{align}
    \mathcal{F}=&-\frac{ND}{2}\log2\pi-\frac{D}{2}\log |\bm{\mathrm{K}}_{nn}+\beta^{-1}\bm{\mathrm{I}}_{n}|\notag\\
    &-\frac{1}{2}\mathrm{tr}\left[(\bm{\mathrm{K}}_{nn}+\beta^{-1}\bm{\mathrm{I}}_{n})^{-1}\bm{\mathrm{Y}}\bm{\mathrm{Y}}^{\top}\right],\label{eq:GP-LVM}    
\end{align}
where $\bm{\mathrm{K}}_{nn}\in\mathbb{R}^{N \times N}$ is a gram matrix whose $(i,\,j)$-th entry is $k(\bm{\mathrm{x}}_{i},\bm{\mathrm{x}}_{j})$. The maximization of Eq.~\eqref{eq:GP-LVM} is performed through the gradient-based optimization; however, the evaluation of Eq.~\eqref{eq:GP-LVM} requires cubic time complexity $O(N^{3})$, which restricts the scalability of the GP-LVM. In addition, the nonlinearity of $(\bm{\mathrm{K}}_{nn}+\beta^{-1}\bm{\mathrm{I}}_{n})^{-1}$ hinders the full Bayesian treatment of $\bm{\mathrm{X}}$.
The \textit{inducing points method}~\citep{titsias2009variational, bauer2016understanding} has been used to address these issues. This method assumes inducing point $\bm{\mathrm{u}}_{d}$, and its positions $\bm{\mathrm{z}}_{k}\in\mathcal{M}^{Q}\,(k=1,2,\ldots,M)$ are \textit{sufficient statistics} for the prior $\bm{\mathrm{f}}_{d}$, i.e., $p(\bm{\mathrm{Y}}|\bm{\mathrm{X}}, \bm{\mathrm{Z}})=\prod_{d=1}^{D}\int p(\bm{\mathrm{y}}_{:,d}|\bm{\mathrm{u}}_{d},\bm{\mathrm{X}},\bm{\mathrm{Z}})p(\bm{\mathrm{u}}_{d}|\bm{\mathrm{Z}})d\bm{\mathrm{u}}_{d}$, and then introduces the variational inference as $\log p(\bm{\mathrm{y}}_{:,d}|\bm{\mathrm{u}}_{d},\bm{\mathrm{X}},\bm{\mathrm{Z}})\ge\mathbb{E}_{p(\bm{\mathrm{f}}_{d}|\bm{\mathrm{u}}_{d},\bm{\mathrm{X}},\bm{\mathrm{Z}})}\left[\log p(\bm{\mathrm{y}}_{:,d}|\bm{\mathrm{f}}_{d})\right]$.
The objective function of the sparse GP-LVM is a tight lower bound of the log-likelihood $\sum_{d=1}^{D}\log p(\bm{\mathrm{y}}_{:,d}|\bm{\mathrm{X}},\bm{\mathrm{Z}})\ge\acute{\mathcal{F}}$:
\begin{align}
    \acute{\mathcal{F}}=&-\frac{D}{2}\log\frac{(2\pi)^{N}|\bm{\mathrm{A}}|}{\beta^{N}|\bm{\mathrm{K}}_{mm}|}-\frac{1}{2}\mathrm{tr}\left(\bm{\mathrm{W}}\bm{\mathrm{Y}}\bm{\mathrm{Y}}^{\top}\right)\notag\\
    &-\frac{\beta D}{2}\mathrm{tr}\left(\bm{\mathrm{K}}_{nn}\right)+\frac{\beta D}{2}\left(\bm{\mathrm{K}}_{mm}^{-1}\bm{\mathrm{K}}_{mn}\bm{\mathrm{K}}_{nm}\right),\label{eq:sparse_GP-LVM}
\end{align}
where $\bm{\mathrm{A}}=\bm{\mathrm{K}}_{mm}+\beta\bm{\mathrm{K}}_{mn}\bm{\mathrm{K}}_{nm}$, $\bm{\mathrm{W}}=\beta\bm{\mathrm{I}}_{n}-\beta^{2}\bm{\mathrm{K}}_{nm}\bm{\mathrm{A}}^{-1}\bm{\mathrm{K}}_{mn}$, and $\bm{\mathrm{K}}_{mn}=\bm{\mathrm{K}}_{nm}^{\top}\in\mathbb{R}^{M \times N}$ are gram matrices whose $(k,\,j)$-th entry is $k(\bm{\mathrm{z}}_{k},\bm{\mathrm{x}}_{j})$. The problems regarding the scalability and nonlinearity of Eq.~\eqref{eq:GP-LVM} are solved by vanishing the inversion of $\bm{\mathrm{K}}_{nn}+\beta^{-1}\bm{\mathrm{I}}_{n}$. The Bayesian GP-LVM estimates the approximated posterior $q(\bm{\mathrm{X}})=\prod_{i=1}^{N}q(\bm{\mathrm{x}}_{i})$ rather than the deterministic latent variables. The objective is the evidence lower bound (ELBO) $\sum_{d=1}^{D}\log p(\bm{\mathrm{y}}_{:,d}|\bm{\mathrm{Z}})\ge\acute{\mathcal{F}}_{b}$, which is expressed as follows:
\begin{align}
    \acute{\mathcal{F}}_{b}&=-\frac{D}{2}\log\frac{(2\pi)^{N}|\bm{\mathrm{A}}_{b}|}{\beta^{N}|\bm{\mathrm{K}}_{mm}|}-\frac{1}{2}\mathrm{tr}\left(\bm{\mathrm{W}}_{b}\bm{\mathrm{Y}}\bm{\mathrm{Y}}^{\top}\right)\notag\\
    &-\frac{\beta D}{2}\mathrm{tr}\left(\bm{\mathrm{K}}_{nn}\right)+\frac{\beta D}{2}\mathrm{tr}\left(\bm{\mathrm{K}}_{mm}^{-1}\bm{\Psi}_{2}\right)-\sum_{i=1}^{N}\mathrm{KL}_{i},\label{eq:Bayesian_sparse_GP-LVM}
\end{align}
where $\mathrm{KL}_{i}=\mathrm{KL}[q(\bm{\mathrm{x}}_{i})||p(\bm{\mathrm{x}}_{i})]$ is Kullback–Leibler (KL) divergence between $q(\bm{\mathrm{x}}_{i})$ and $p(\bm{\mathrm{x}}_{i})$, $p(\bm{\mathrm{x}}_{i})$ is a prior distribution of latent variables, $\bm{\mathrm{A}}_{b}=\bm{\mathrm{K}}_{mm}+\beta\bm{\Psi}_{2}$, $\bm{\mathrm{W}}_{b}=\beta\bm{\mathrm{I}}_{n}-\beta^{2}\bm{\mathrm{\Psi}}_{1}^{\top}\bm{\mathrm{A}}_{b}^{-1}\bm{\mathrm{\Psi}}_{1}$, $\bm{\Psi}_{1}=\mathbb{E}_{q(\bm{\mathrm{X}})}[\bm{\mathrm{K}}_{mn}]$, and $\bm{\Psi}_{2}=\mathbb{E}_{q(\bm{\mathrm{X}})}[\bm{\mathrm{K}}_{mn}\bm{\mathrm{K}}_{nm}]$. Note that we provide the detailed derivation from Eq.~\eqref{eq:GP-LVM} to Eq.~\eqref{eq:Bayesian_sparse_GP-LVM} in Appendix A.1.
The computational intractability of Eq.~\eqref{eq:Bayesian_sparse_GP-LVM} appears frequently in the marginalization of $\bm{\Psi}$ statistics, which is typically solved by sampling approximation with the reparameterization trick~\citep{salimbeni2017doubly, de2021learning, lalchand2022generalised}.

\subsection{Hyperbolic Geometry}
\label{section:hyperbolic_space}
Hyperbolic spaces are smooth Riemannian manifolds and have several isometric models~\citep{peng2021hyperbolic}. For example, the \textit{Poincar{\'e} ball model}~\citep{nickel2017poincare, ganea2018hyperbolic} $\mathcal{P}^{Q}=(\mathbb{B}^{Q},g_{b})$ is a representative example in machine learning literature, where $\mathbb{B}^{Q}=\{\bm{\mathrm{x}}\in\mathbb{R}^{Q}:||\bm{\mathrm{x}}||_{2}<1\}$, and $g_{b}=\frac{2}{1-||\bm{\mathrm{x}}||^{2}}g_{e}$ is the metric tensor with the Euclidean metric tensor $g_{e}$. However, the boundary $\{\bm{\mathrm{x}}\in\mathbb{R}^{Q}:||\bm{\mathrm{x}}||_{2}=1\}$ in the Poincar{\'e} ball model causes numerical instability. The \textit{Lorentz model} is another example that does not contain a boundary. Formally, the Lorentz model $\mathcal{L}^{Q}=(\mathbb{H}^{Q}, g_{l})$ is a $Q$-dimensional hyperbolic space with a $Q$-dimensional \textit{upper hyperboloid} $\mathbb{H}^{Q}=\{\bm{\mathrm{x}}=[x_{0},x_{1},\ldots,x_{Q}]^{\top}\in\mathbb{R}^{Q+1}:\,\left<\bm{\mathrm{x}},\bm{\mathrm{x}}\right>_{\mathcal{L}^{Q}}=-1,\,x_{0}>0\}$ and metric tensor $g_{l}=\mathrm{diag}(-1,1,\ldots,1)\in\mathbb{R}^{(Q+1)\times(Q+1)}$, where $\left<\bm{\mathrm{x}},\bm{\mathrm{x}}'\right>_{\mathcal{L}^{Q}}=-x_{0}x'_{0}+\sum_{i=1}^{Q}x_{i}x'_{i}$ is the \textit{Lorentzian inner product}. Note that $x_{0}=\sqrt{1+||\tilde{\bm{\mathrm{x}}}||_{2}^{2}}$, where $\tilde{\bm{\mathrm{x}}}=[x_{1},x_{2},\ldots,x_{Q}]^{\top}$. The tangent space $\mathcal{T}_{\bm{\mu}}\mathcal{L}^{Q}$, i.e., the set of the tangent passes through $\bm{\mu}\in\mathcal{L}^{Q}$, is useful to extend Euclidean operations to the hyperbolic space. Furthermore, the mapping from the Lorentz model onto its tangent space at $\bm{\mu}$ is stated explicitly by the exponential map $\mathrm{Exp}_{\bm{\mu}}(\bm{\mathrm{v}}):\mathcal{T}_{\bm{\mu}}\mathcal{L}^{Q}\rightarrow\mathcal{L}^{Q}$ as follows:
\begin{align}
    \mathrm{Exp}_{\bm{\mu}}(\bm{\mathrm{v}})&=\cosh(||\bm{\mathrm{v}}||_{\mathcal{L}^{Q}})\bm{\mu}+\sinh(||\bm{\mathrm{v}}||_{\mathcal{L}^{Q}})\frac{\bm{\mathrm{v}}}{||\bm{\mathrm{v}}||_{\mathcal{L}^{Q}}},\label{eq:exp_map}
\end{align}
where $||\bm{\mathrm{v}}||_{\mathcal{L}^{Q}}=\sqrt{\left<\bm{\mathrm{v}},\bm{\mathrm{v}}\right>_{\mathcal{L}^{Q}}}$ is the norm of $\bm{\mathrm{v}}\in\mathcal{T}_{\bm{\mu}}\mathcal{L}^{Q}$. The length of geodesic between two points on $\mathcal{L}^{Q}$ is expressed as follows:
\begin{align}
    d_{\mathcal{L}^{Q}}(\bm{\mathrm{x}},\bm{\mathrm{x}}')=\cosh^{-1}\left(-\left<\bm{\mathrm{x}},\bm{\mathrm{x}}'\right>_{\mathcal{L}^{Q}}\right).\label{eq:dist}
\end{align}
Finally, the diffeomorphism $p(\bm{\mathrm{x}}):\mathcal{L}^{Q}\rightarrow\mathcal{P}^{Q}$ is given as:
\begin{align} 
  p(\bm{\mathrm{x}})=\frac{[x_{1},\,x_{2},\,\ldots,x_{Q}]^{\top}}{1+x_{0}}.\label{eq:diffeomorphism} 
\end{align}
We learn the latent variables on the Lorentz model and visualize them on the Poincar{\'e} ball model.

\section{Hyperboloid GP-LVMs}
The proposed hGP-LVMs, which are the primary contribution of this paper, and presented in the following. First, we establish the positive definite (PD) kernel on the Lorentz model, which is referred to as the \textit{hyperboloid exponential kernel}. We then explain the optimization of the hGP-LVMs considering the curved geometry of the latent space. In the Bayesian estimation method, straightforward optimization cannot be realized due to the computational intractability included in the $\bm{\Psi}$ statistics and KL divergence. Thus, we develop a dedicated algorithm that combines the reparameterization trick~\citep{kingma2013auto}, Riemannian optimization~\citep{nickel2018learning}, and active set approximation of GP-LVMs~\citep{moreno2022revisiting}.
\subsection{Hyperboloid Exponential Kernel}
\label{subsec:hyperboloid_kernel}
\begin{figure}
    \centering
    {
        \includegraphics[width=70mm]{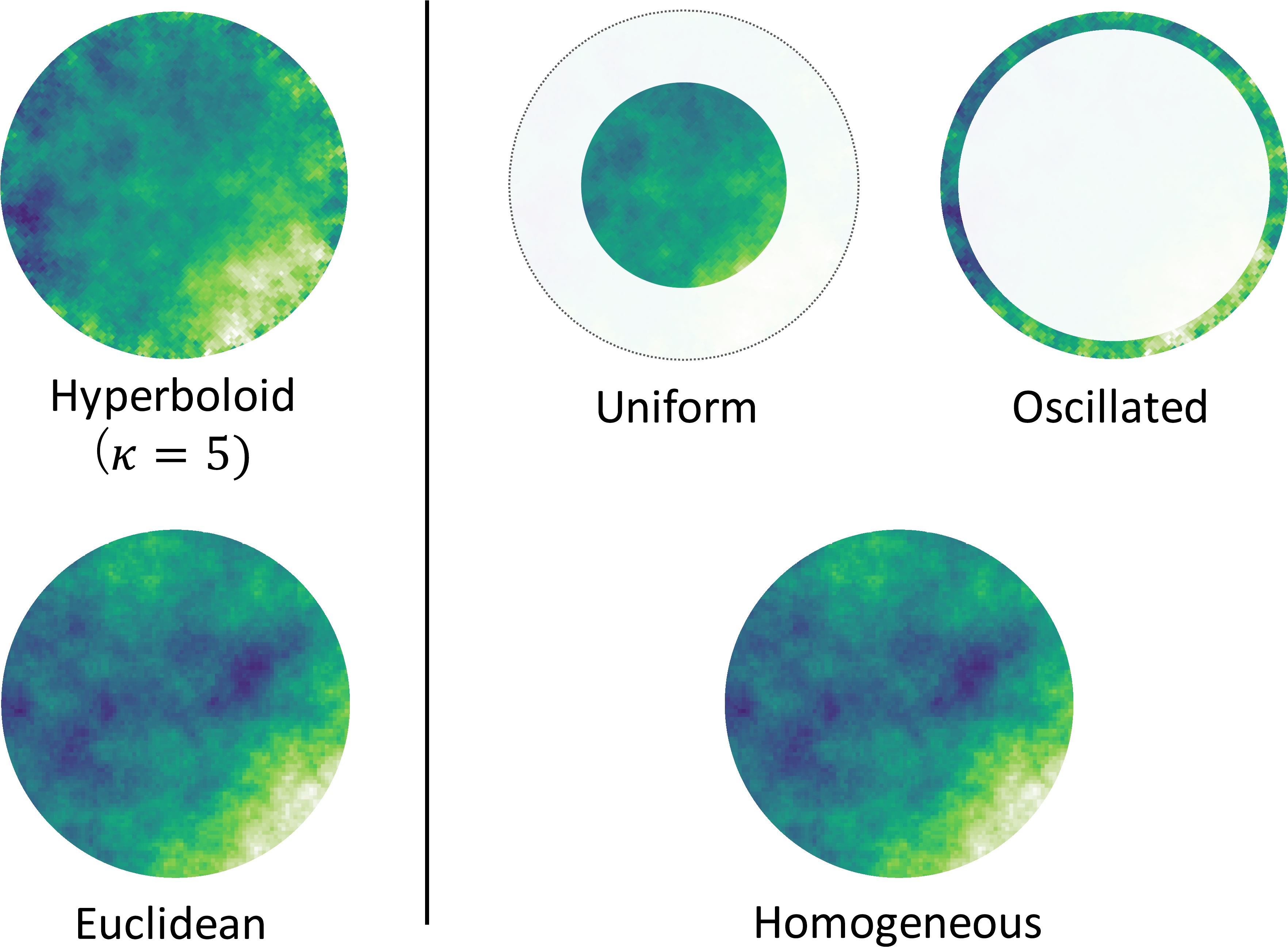}
    }
    \caption{GP prior comparison between the hyperboloid exponential kernel (\textit{upper}, $\kappa=5$) and the Euclidean exponential kernel (\textit{bottom}). The color gives the value of the sampled GP. We input the latent variables on the Poincar\'{e} ball model when $\mathcal{M}=\mathcal{L}^{Q}$ (\textit{left}) and those on the unit circle when $\mathcal{M}=E$ (\textit{right}).}
    \label{fig:GP_sampling}
    \vspace{-0.3cm}
\end{figure}
Recent studies have utilized the heat kernel for the Riemannian kernel construction~\citep{atigh2022hyperbolic, niu2023intrinsic, azangulov2023stationary}; however, it has a high computational cost, which hinders the scalable optimization. Thus, we employ the result reported in the literature~\citep{feragen2015geodesic} that shows that the \textit{geodesic exponential kernel} is the PD kernel under a certain condition. Formally, the geodesic exponential kernel is expressed as follows:
\begin{align}
    k_{\mathcal{M}}(\bm{\mathrm{x}},\bm{\mathrm{x}}')=\sigma\exp\left(-\frac{d_{\mathcal{M}}(\bm{\mathrm{x}},\bm{\mathrm{x}}')}{\kappa}\right),\label{eq:geodesic_exponential_kernel}
\end{align}
where $\sigma$ and $\kappa$ are variance and \textit{length scale} parameters,respectively. This geodesic exponential kernel can be PD if $d_{\mathcal{M}}(\bm{\mathrm{x}},\bm{\mathrm{x}}')$ is a conditionally negative definite (CND) metric~\citep{feragen2015geodesic}. Here, the hyperbolic metric is CND~\citep{istas2012manifold}; thus, we can apply the geodesic exponential kernel to the hyperbolic space. When $\mathcal{M}=\mathcal{L}^{Q}$, we refer to the kernel function as the \textit{hyperboloid exponential (HE) kernel}. To visualize the difference between the HE and Euclidean kernels, Figure~\ref{fig:GP_sampling} compares the GP prior of the HE kernel and that of the Euclidean exponential kernel. When $\mathcal{M}=\mathcal{L}^{Q}$, the prior highly correlates around the origin (similar values) and oscillates around the rim (striped values), unlike the Euclidean kernel, which is homogeneous on the entire unit circle. This heterogeneity of the hyperboloid kernel contributes to embedding the hierarchical data containing the exponentially growing child nodes.
\par
In the following, we discuss the role of the HE kernel parameters $\sigma$ and $\kappa$. The variance $\sigma$ is similar to the general exponential kernel, i.e., it determines the range of the kernel function. However, the \textit{length scale} parameter $\kappa$ differs from that of the Euclidean kernel. The length scale of Euclidean GP-LVMs calibrates the scale of the latent variables. In the flat Euclidean space, the scale has no structural meaning; however, it does not hold in the curved hyperbolic space. With a large length scale, the latent variables are widely spread on the curved manifold and are strongly influenced by the hyperbolic curvature. With a small length scale, we first show one result for the relation between the distance in the Lorentz model and the Euclidean space.
\begin{lemma}
Here, let $\bm{\mathrm{x}}_{
1}\in\mathbb{R}^{Q}$ and $\bm{\mathrm{x}}_{2}\in\mathbb{R}^{Q}$ be vectors  with small scales, i.e., $||\bm{\mathrm{x}}_{1}||_{2}\approx0,\,||\bm{\mathrm{x}}_{2}||_{2}\approx0$, and set $\bm{\mathrm{x}}_{1}'=\left[\sqrt{1+||\bm{\mathrm{x}}_{1}||_{2}^{2}},\,
\bm{\mathrm{x}}_{1}^{\top}\right]^{\top}\in\mathcal{L}^{Q}$ and $\bm{\mathrm{x}}_{2}'=\left[\sqrt{1+||\bm{\mathrm{x}}_{2}||_{2}^{2}},\,\bm{\mathrm{x}}_{2}^{\top}\right]^{\top}\in\mathcal{L}^{Q}$. Then, we obtain the following result:
\begin{align}
    d_{\mathcal{L}^{Q}}(\bm{\mathrm{x}}_{1}',\,\bm{\mathrm{x}}_{2}')\approx d_{E}(\bm{\mathrm{x}}_{1},\bm{\mathrm{x}}_{2})+O(d_{E}(\bm{\mathrm{x}}_{1},\bm{\mathrm{x}}_{2})^{3})\label{eq:distance_relationship},
\end{align}
\textit{where $d_{E}(\bm{\mathrm{x}}_{1},\,\bm{\mathrm{x}}_{2})=||\bm{\mathrm{x}}_{1}-\bm{\mathrm{x}}_{2}||_{2}$.}
\label{lemma1}
\end{lemma}
\begin{proof}
    Refer to Appendix A.2.
\end{proof}
Equation \eqref{eq:distance_relationship} indicates that the hyperboloid metric approaches the Euclidean metric around the origin asymptotically, and the HE kernel with a small length scale behaves like the Euclidean geodesic kernel. From the above discussion, we understand that \textit{the meaning of the length scale $\kappa$ is how much we expect the latent variables to follow the hyperbolic curvature}. However, this metamorphosis of the length scale causes difficulty with optimization. Thus, we treat the length scale as a predefined hyperparameter and determine it experimentally. In summary, the HE kernel is established by extending the geodesic exponential kernel in Eq.~\eqref{eq:geodesic_exponential_kernel}, and the length scale is treated as a hyperparameter to determine the degree to which the latent variables follow the hyperbolic curvature.

\subsection{Optimization}
Here, we discuss applying the objective of the GP-LVM, sparse GP-LVM, and Bayesian GP-LVM with the hyperboloid kernel to learn the latent variables on the Lorentz model. The optimization challenge is the gradient computation of the latent variables, and we must consider the curved geometry of the hyperboloid. First, we explain the optimization of the deterministic latent variables and extend it to the Bayesian case using a \textit{wrapped Gaussian distribution}~\citep{nagano2019wrapped, mathieu2019continuous, cho2022rotated}.
\begin{algorithm}[t]
\caption{Updating point latent variables.}
\label{alg1}
\begin{algorithmic}[1]
  \REQUIRE learning rate $\alpha$
  \WHILE{$t < \mathsf{max\_iter}$}
  \FOR{$i=1,2,\ldots,N$}
  \STATE $\bm{\mathrm{g}}_{i}^{t}\gets g_{l}^{-1}[\frac{\partial \mathcal{F}}{\partial x_{i0}^{t}},\,\frac{\partial \mathcal{F}}{\partial x_{i1}^{t}},\,\ldots,\frac{\partial \mathcal{F}}{\partial x_{iQ}^{t}}]^{\top}$
  \STATE $\mathcal{T}\bm{\mathrm{g}}_{i}^{t}\gets\mathrm{proj}_{\bm{\mathrm{x}}_{i}^{t}}(\bm{\mathrm{g}}_{i}^{t})$
  \STATE $\bm{\mathrm{x}}_{i}^{t+1}\gets\exp_{\bm{\mathrm{x}}_{i}^{t}}(-\alpha\mathcal{T}\bm{\mathrm{g}}_{i}^{t})$
  \ENDFOR
\ENDWHILE
\end{algorithmic}
\end{algorithm}
\paragraph{hGP-LVM and Sparse hGP-LVM.}
\label{section:Methods}
We consider the curved surfaces of the latent space by employing the Riemannian gradient descent algorithm~\citep{nickel2018learning}. This algorithm (\rom{1}) computes the steepest direction, (\rom{2}) projects the \textit{row} gradients into the tangent space, and (\rom{3}) wraps the projected vector in the surface of the Lorentz model following Eq.~\eqref{eq:exp_map} (\textbf{Algorithm}~\ref{alg1}). Here, the row gradients are given as $\frac{\partial\mathcal{F}}{\partial x_{iq}}=\mathrm{tr}(\frac{\partial\mathcal{F}}{\partial\bm{\mathrm{K}}}\frac{\partial\bm{\mathrm{K}}}{\partial x_{iq}})$. In addition, the projection from the ambient Euclidean space to the Lorentz model $\mathrm{proj}_{\bm{\mu}}(\bm{\mathrm{g}}):\mathbb{R}^{N+1}\rightarrow\mathcal{T}_{\bm{\mu}}\mathcal{L}^{Q}$ is expressed as follows:
\begin{align}
    \mathrm{proj}_{\bm{\mu}}(\bm{\mathrm{g}})=\bm{\mathrm{g}}+\left<\bm{\mu},\bm{\mathrm{g}}\right>_{\mathcal{L}^{Q}}\bm{\mu}.\label{eq:proj_euc_to_TuL}
\end{align}
In the sparse hGP-LVM, the positions of the inducing points $\bm{\mathrm{Z}}$ must be optimized in addition to the latent variables; however, their gradient-based updates cause unstable optimization (even in the Euclidean case). Therefore, we employ the active set approximation scheme~\citep{moreno2022revisiting}, where we sample the inducing positions $\bm{\mathrm{Z}}$ from the latent variables $\bm{\mathrm{X}}$ at each multiple updates.
\paragraph{Bayesian hGP-LVM.}
To establish probabilistic latent variable models, we first require the variational distribution $q(\bm{\mathrm{x}}_{i})$ and prior $p(\bm{\mathrm{x}}_{i})$ defined
on the Lorentz models. Here, for computational efficiency we employ the wrapped Gaussian distribution $\mathcal{N}_{\mathcal{L}^{Q}}^{w}(\bm{\mathrm{x}}|\bm{\mu},\,\bm{\mathrm{S}})$~\citep{nagano2019wrapped} rather than the exact hyperbolic Gaussian. In addition, we assume $q(\bm{\mathrm{x}}_{i})=\mathcal{N}_{\mathcal{L}^{Q}}^{w}(\bm{\mathrm{x}}_{i}|\bm{\mathrm{\mu}}_{i},\,\bm{\mathrm{S}}_{i})$ and $p(\bm{\mathrm{x}}_{i})=\mathcal{N}_{\mathcal{L}^{Q}}^{w}(\bm{\mathrm{x}}_{i}|\bm{\mathrm{0}},\,\bm{\mathrm{I}}_{q})$, where $\bm{\mathrm{\mu}}_{i}\in\mathcal{L}^{Q}$ and $\bm{\mathrm{S}}_{i}=\mathrm{diag}(s_{i1},\,s_{i2},\,\ldots,s_{iQ})\in\mathbb{R}^{Q \times Q}$ are variational parameters. The probability density function of $\mathcal{N}_{\mathcal{L}^{Q}}^{w}(\bm{\mathrm{x}}_{i}|\bm{\mathrm{\mu}}_{i},\,\bm{\mathrm{S}}_{i})$ can be given in a closed form as follows:
\begin{align}
    \mathcal{N}_{\mathcal{L}^{Q}}^{w}(\bm{\mathrm{x}}|\bm{\mu},\bm{\mathrm{S}})&=\left\{\frac{\sinh(||\bm{\mathrm{u}}||_{\mathcal{L}^{Q}})}{||\bm{\mathrm{u}}||_{\mathcal{L}^{Q}}}\right\}^{(Q-1)}\mathcal{N}(\bm{\mathrm{v}}|\bm{0},\bm{\mathrm{S}}).\label{eq:wrapped_gaussian_distribution}
\end{align}
Computing the $\bm{\Psi}$ statistics and KL divergence in Eq.~\eqref{eq:Bayesian_sparse_GP-LVM} is intractable with the HE kernel; thus, we compute them approximately using the sampling scheme and the reparameterization trick. The sampling scheme of $\mathcal{N}_{\mathcal{L}^{Q}}^{w}(\bm{\mathrm{x}}|\bm{\mu},\,\bm{\mathrm{S}})$  involves three steps (similar to the Riemannian optimization), i.e., (\rom{1}) sampling $\widetilde{\bm{\mathrm{v}}}\sim\mathcal{N}(\bm{0},\,\bm{\mathrm{S}})$ at the origin $\bm{\mu}_{0}=[1,0,\ldots,0]^{\top}\in\mathcal{L}^{Q}$, (\rom{2}) carrying $\bm{\mathrm{v}}=[0,\widetilde{\bm{\mathrm{v}}}^{\top}]^{\top}\in\mathcal{T}_{\bm{\mu}_{0}}\mathcal{L}^{Q}$ from the origin to an arbitrary point $\bm{\mu}$ as $\bm{\mathrm{u}}=\mathrm{PT}_{\bm{\mu}_{0}\rightarrow\bm{\mu}}(\bm{\mathrm{v}})$, and (\rom{3}) wrapping the vector in the surface following Eq.~\eqref{eq:exp_map}. The parallel transportation $\mathrm{PT}_{\bm{\nu}\rightarrow\bm{\mu}}(\bm{\mathrm{v}}):\mathcal{T}_
{\bm{\nu}}\mathcal{L}^{Q}\rightarrow\mathcal{T}_
{\bm{\mu}}\mathcal{L}^{Q}$ is computed as follows:
\begin{align}
    \mathrm{PT}_{\bm{\nu}\rightarrow\bm{\mu}}(\bm{\mathrm{v}})=\bm{\mathrm{v}}+\frac{\left<\bm{\mu}-\gamma\bm{\nu},\bm{\mathrm{v}}\right>_{\mathcal{L}^{Q}}}{\gamma+1}(\bm{\mu}+\bm{\nu}),\label{eq:parallel_transport}
\end{align}
where $\gamma=\left<\bm{\mu},\,\bm{\nu}\right>_{\mathcal{L}^{Q}}$.
Thus, the Monte Carlo approximations of $\bm{\Psi}$ and the KL divergence are obtained as follows: 
\begin{align}
    [\bm{\Psi}_{1}]_{ik}&=\sum_{h=1}^{H}k_{\mathcal{L}^{Q}}(\bm{\mathrm{x}}_{i}^{(h)},\bm{\mathrm{z}}_{k}),\,\bm{\Psi}_{2}=\sum_{i=1}^{N}\bm{\Psi}_{2}^{(i)},\label{eq:psi_1_and_psi_2}\\
    [\bm{\Psi}_{2}^{(i)}]_{kl}&=\sum_{h=1}^{H}k_{\mathcal{L}^{Q}}(\bm{\mathrm{z}}_{k},\bm{\mathrm{x}}_{i}^{(h)})k_{\mathcal{L}^{Q}}(\bm{\mathrm{x}}_{i}^{(h)},\bm{\mathrm{z}}_{l}),\label{eq:Monte_carlo_Psi_statistics}\\
    \mathrm{KL}_{i}&=\sum_{h=1}^{H}\log\frac{\mathcal{N}_{\mathcal{L}^{Q}}^{w}(\bm{\mathrm{x}}_{i}^{(h)}|\bm{\mu}_{i},\bm{\mathrm{S}}_{i})}{\mathcal{N}_{\mathcal{L}^{Q}}^{w}(\bm{\mathrm{x}}_{i}^{(h)}|\bm{0},\bm{\mathrm{I}}_{q})}.\label{eq:Monte_carlo_KL_div}
\end{align}
Here, $H$ is the number of samples, and $\bm{\mathrm{x}}_{i}^{(h)}$ is the reparameterized latent variables:
\begin{align}
    \bm{\mathrm{x}}_{i}^{(h)}&=\mathrm{Exp}_{\bm{\mu}_{i}}(\bm{\mathrm{u}}_{i}^{(h)}),\\
    \bm{\mathrm{u}}_{i}^{(h)}&=\mathrm{PT}_{\bm{\mu}_{0}\rightarrow\bm{\mu}_{i}}(\bm{\mathrm{v}}_{i}^{(h)}),\\
    \tilde{\bm{\mathrm{v}}}_{i}^{(h)}&=\bm{\mathrm{S}}_{i}^{\frac{1}{2}}\bm{\zeta}_{i}^{(h)},\label{eq:sampling}
\end{align}
where $\bm{\mathrm{v}}_{i}=[0,\widetilde{\bm{\mathrm{v}}}_{i}^{(h)\top}]^{\top}$ and $\bm{\zeta}_{i}^{(h)}\sim\mathcal{N}(\bm{0},\bm{\mathrm{I}}_{q})$. Note that we update the variational mean $\bm{\mathrm{\mu}}_{i}$ following the Riemannian procedure described in \textbf{Algorithm}~\ref{alg1}, and we determine $\bm{\mathrm{Z}}$ by sampling from $\bm{\mu}_{i}$. The computation of the row gradient $\frac{\partial \acute{\mathcal{F}}_{b}}{\partial \bm{\mu}_{i}}$ and $\frac{\partial \acute{\mathcal{F}}_{b}}{\partial \bm{\mathrm{S}}_{i}}$ is described in detail in Appendix A.3. 
\par
The time complexities of hGP-LVM, sparse hGP-LVM, and Bayesian hGP-LVM are $O(N^{3})$, $O(M^{2}N+M^{3})$, and $O(HM^{2}N+HM^{3})$, respectively, which are equal to the costs of the previous GP-LVM baselines.

\section{Related Work}
\label{section:Related_Works}

\paragraph{Visualization-Aided DR.} t-SNE~\citep{van2008visualizing} is the gold-standard method for data visualization, and the recent UMAP~\citep{mcinnes2018umap} approach introduces the algebraic topology concept and realizes DR with a solid theoretical foundation. Note that the neighbor embedding methods frequently overlook the global structure; thus, recent variants attempt to catch the global structure using a triplet loss function~\citep{amid2019trimap} or by modifying the initial embedding~\citep{kobak2019art, wang2021understanding}. DR for hierarchical data has gained considerable attention, where the potential heat diffusion for affinity-based trajectory embedding (PHATE)~\citep{moon2019visualizing} method employs the diffusion operation, and the Poincar\'{e}Map~\citep{klimovskaia2020poincare} develops the neighbor embedding on the Poincar\'{e} ball model. Visualization-aided DR methods have been widely applied in scRNA-seq analysis~\citep{luecken2019current} to support the exposition of complex cell biology.

\paragraph{Hyperbolic Machine Learning.} Previously, the hyperbolic space has been employed to realize the hierarchical structure learning of word taxonomies~\citep{nickel2017poincare, nickel2018learning}. Recent studies have applied hyperbolic neural networks to various other tasks, including image segmentation~\citep{atigh2022hyperbolic}, knowledge graph embedding~\citep{chami2020low}, or image-text representation learning~\citep{desai2023hyperbolic}. Gyrovector space equipped M\"{o}bius addition and scalar multiplication~\citep{ganea2018hyperbolic} have also been investigated to define the layer-to-layer transformation of fully connected hyperbolic networks. For unsupervised learning, VAEs with a wrapped Gaussian distribution have also been studied extensively~\citep{nagano2019wrapped,mathieu2019continuous, cho2022rotated}, and hyperbolic kernel methods have been explored for support vector machines~\citep{fan2023horospherical}. In addition, several PD kernels on the Poincar\'{e} ball have been presented in the literature~\citep{peng2021hyperbolic, yang2023expanding}. 
\begin{figure}
    \centering
    {
        \includegraphics[width=70mm]{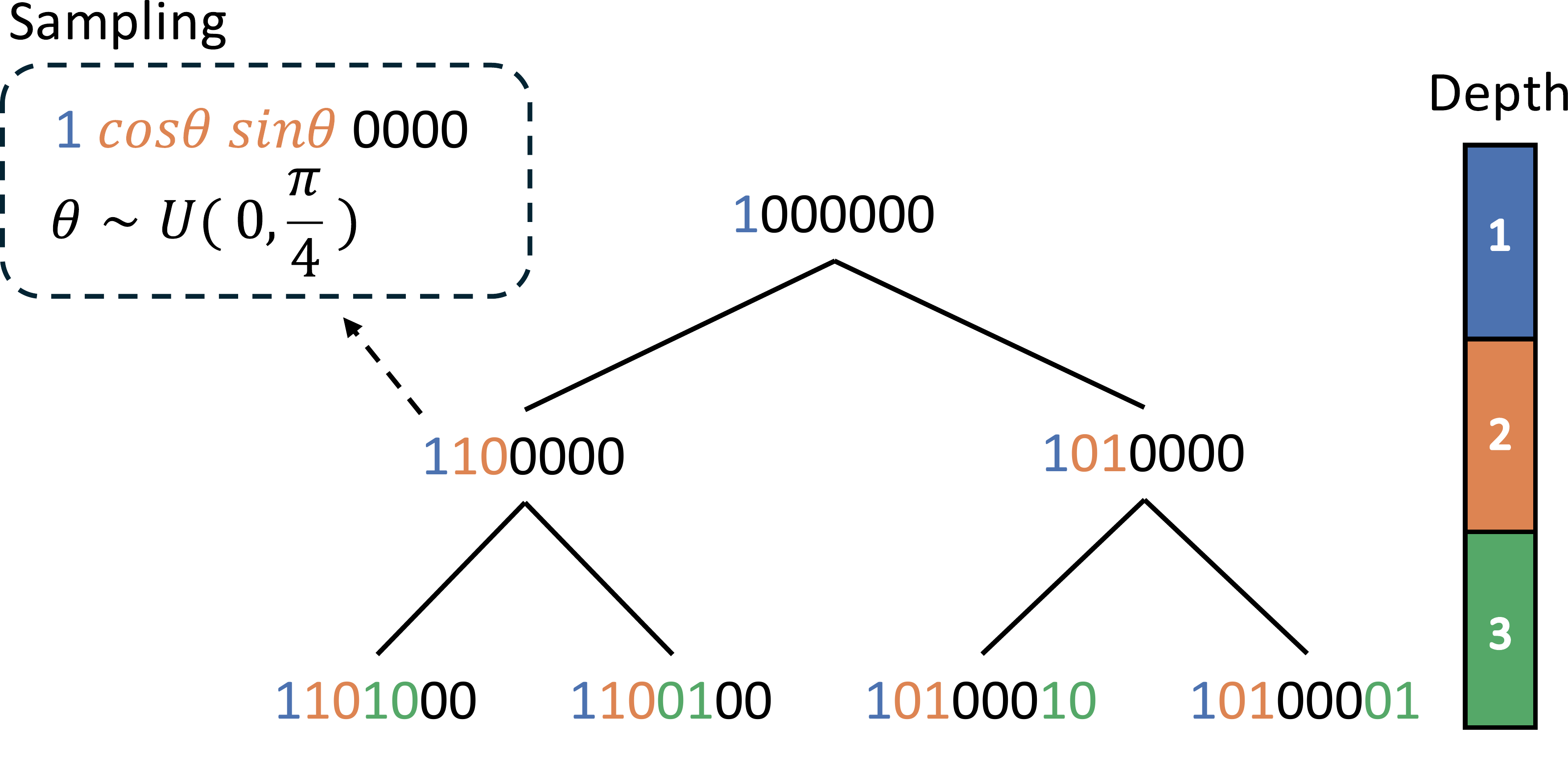}
    }
    \caption{Illustration of the synthetic binary tree (SBT) with $d=3$ and the sampling procedure.}
    \label{fig:main_fig_sbt}
    \vspace{-0.3cm}
\end{figure}

\paragraph{Riemannian GP-LVM.} The Mat\'{e}rn covariance on several curved manifolds has been studied previously~\citep{borovitskiy2020matern}, and the heat kernel in the Lie group was developed~\citep{azangulov2022stationary, azangulov2023stationary}. To the best of our knowledge, only one hyperbolic extension of GP-LVMs with the hyperbolic heat kernel exists for motion taxonomy embedding in the latent space~\citep{jaquier2022bringing}. Its objective includes sampling computation in the hyperbolic heat kernel and $\bm{\Psi}$ computation. This involves high time complexity by double sampling, and this method is intractable for general DR with several hundreds of data points. Furthermore, the manifold GP-LVM (mGP-LVM)~\citep{jensen2020manifold} employs the geodesic exponential kernel to represent the neural curvature on simple manifolds, e.g., tori, Spheres, and SO(3). However, the mGP-LVM involves mathematical difficulty in the computation of the posterior distribution, which restricts the applicability of the mGP-LVM to other smooth manifolds, e.g., hyperbolic spaces.

\section{Experiments}
\label{section:experiments}
In this section, we present experimental results to validate the proposed method. First, we compare the hGP-LVMs with Euclidean and hyperbolic generative models~\citep{nagano2019wrapped, cho2022rotated}. We discuss visualization-aided DR on the scRNA-seq dataset and confirm the effectiveness of our methods, especially the Bayesian estimation method.
\subsection{Synthetic Binary Tree Dataset}
\label{section: binary_tree}

\begin{table}
\caption{Details of SBT dataset.}
\label{tab:statistical_details_SBT_dataset}
\begin{center}
\begin{tabular}{cccc}\toprule
& & Depth &\\ \cmidrule{2-4}
& $d=4$ & $d=5$ & $d=6$ \\ \midrule
\# samples & 300 & 620 & 1,260 \\ 
\# dimensions & 15 & 31 & 63 \\  \bottomrule
\end{tabular}
\end{center}
\vspace{-0.3cm}
\end{table}

\begin{table*}[t]
\caption{\textbf{Quantitative results obtained on the SBT dataset.} The best results are highlighted in blue. Here, we computed the mean and standard deviation over $10$ runs.}
  \begin{center}
      \begin{tabular}{lccc} \toprule
         & & Depth &\\ \cmidrule{2-4}
          & $d=4$ & $d=5$ & $d=6$ \\ \midrule
         hGP-LVM \textbf{(ours)} & $0.816_{\pm 0.012}$ & \colorbox[rgb]{0.9, 0.9, 1.0}{$0.909_{\pm 0.003}$} & \colorbox[rgb]{0.9, 0.9, 1.0}{$0.849_{\pm 0.004}$} \\
         GP-LVM~\citep{lawrence2005probabilistic} & $0.782_{\pm 0.000}$ & $0.746_{\pm 0.000}$ & $0.717_{\pm 0.000}$ \\
         IsoHVAE~\citep{nagano2019wrapped} & $0.890_{\pm 0.022}$ & $0.867_{\pm 0.025}$ & $0.815_{\pm 0.031}$\\
         RotHVAE~\citep{cho2022rotated} & \colorbox[rgb]{0.9, 0.9, 1.0}{$0.896_{\pm 0.027}$} & $0.872_{\pm 0.019}$ & $0.823_{\pm 0.024}$ \\
        \bottomrule
      \end{tabular}
    \label{tab:fig_syn_tree_all_embedding}
  \end{center}
\end{table*}

\begin{figure*}
    \centering
    {
        \includegraphics[width=160mm]{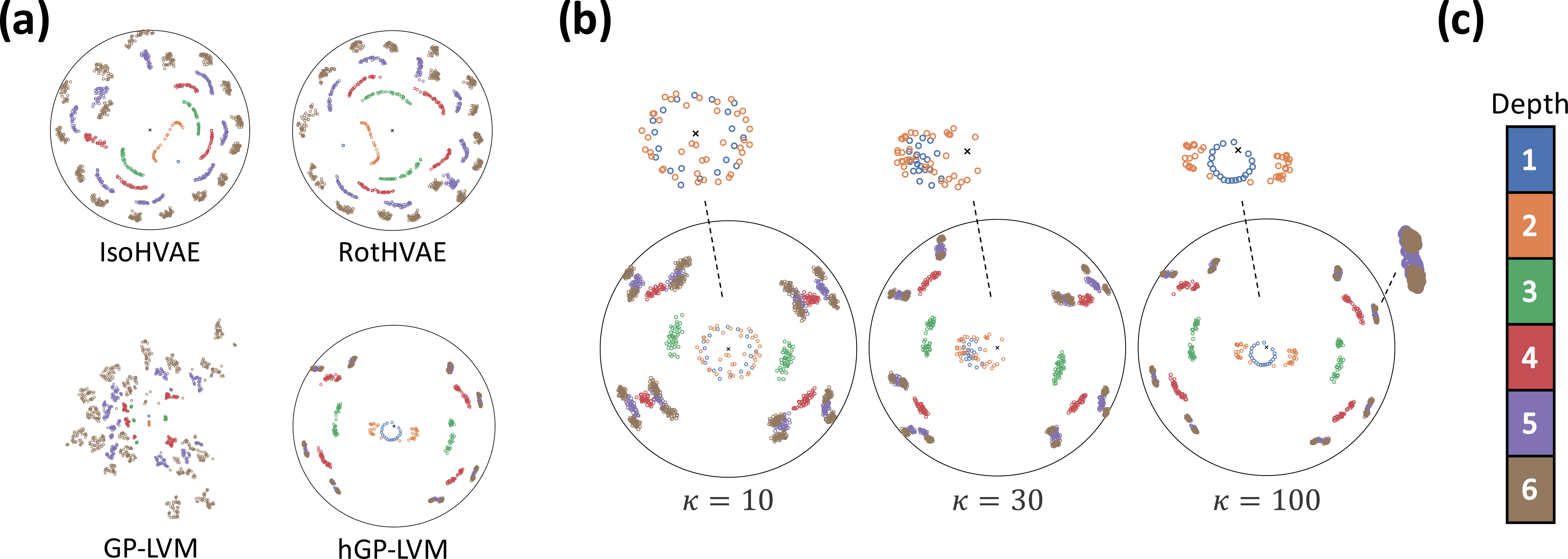}
    }
    \caption{\textbf{Qualitative results obtained on the SBT dataset ($d=6$)}: (a) embedding comparison between generative models, (b) embedding of hGP-LVM with different length scales, and (c) color code of the embedding.}
    \label{fig:main_bin_tree_result}
    \vspace{-0.0cm}
\end{figure*}

In the following, we describe the experiment conducted using the \textit{synthetic binary tree~(SBT) dataset}, which has been used to evaluate hyperbolic generative models. This dataset is based on the SBT, exhibiting a simple hierarchical structure using binary codes. Figure~\ref{fig:main_fig_sbt} shows the SBT and the sampling process of the data points. First, we determine the \textit{depth} of the tree, and then we generate binary codes with lengths of $(2^{d}-1)$. We obtain the code of each node by changing the Boolean values according to the depth. In this study, by oscillating the codes, we generated datasets exhibiting SBT structures. We generated $20$ samples at each node with $d=4,5,6$. The statistical details of these datasets are shown in Table~\ref{tab:statistical_details_SBT_dataset}. Note that we used the most basic hGP-LVM and set $\kappa=100$ in all datasets. In this evaluation, we compared the hGP-LVM with GP-LVM~\citep{lawrence2005probabilistic}, the hyperbolic VAE with an isotropic hyperbolic wrapped Gaussian in Eq.\eqref{eq:wrapped_gaussian_distribution} (IsoHVAE)~\citep{nagano2019wrapped}, and the hyperbolic VAE with a rotated hyperbolic wrapped Gaussian (RotHVAE)~\citep{cho2022rotated}. We also used the implementation and hyperparameter settings presented in the literature~\citep{cho2022rotated} for the hyperbolic VAEs, and we only searched the number of hidden units. The latent dimension of all methods was set to $2$, and we compared the methods through a quantitative evaluation and a qualitative visualization. In the quantitative evaluation, we used the distance correlation score between the latent variables and the observed variables. In this study, the distance between the latent variables was computed using the hyperbolic metric for the hGP-LVM and the hyperbolic VAEs, and the Euclidean metric was computed for the GP-LVM. In addition, the distance between the observed variables was computed with the Hamming distance using the corresponding binary codes to each node. 
\par
The quantitative results are presented in Table~\ref{tab:fig_syn_tree_all_embedding}. As can be seen, the results demonstrate that the hGP-LVM embedded the synthetic hierarchy with higher distance preservation quality than that of the Euclidean GP-LVM on all datasets. The priority of the hGP-LVM over the hyperbolic VAEs was also confirmed for cases where $d=5$ and $d=6$. Although the data size was limited, and more effective parameters may exist for the hyperbolic VAEs, we emphasize that the hGP-LVM did not require such ill-posed parameter tuning and could embed the hierarchy even with a limited amount of data. The visualization comparison is shown in Figure~\ref{fig:main_bin_tree_result}~(a). With the GP-LVM, it is difficult to find the hierarchical nature of the SBT dataset, and the embedding is spread with increasing depth. Although the embedding of the hyperbolic VAEs represented a hierarchical nature, the root nodes with depth $1$ did not position around the origin, and it is difficult to interpret the tree structure in the SBT dataset. In contrast, the proposed hGP-LVM embedded the SBT's hierarchy holding internode similarity and placed the root nodes on the origin, which largely contributed to the visibility of the embedding. In summary, we validated the effectiveness of hGP-LVM toward generative models by conducting an experiment typically used for evaluating hyperbolic generative models.

\begin{figure*}
    \centering
    {
        \includegraphics[width=170mm]{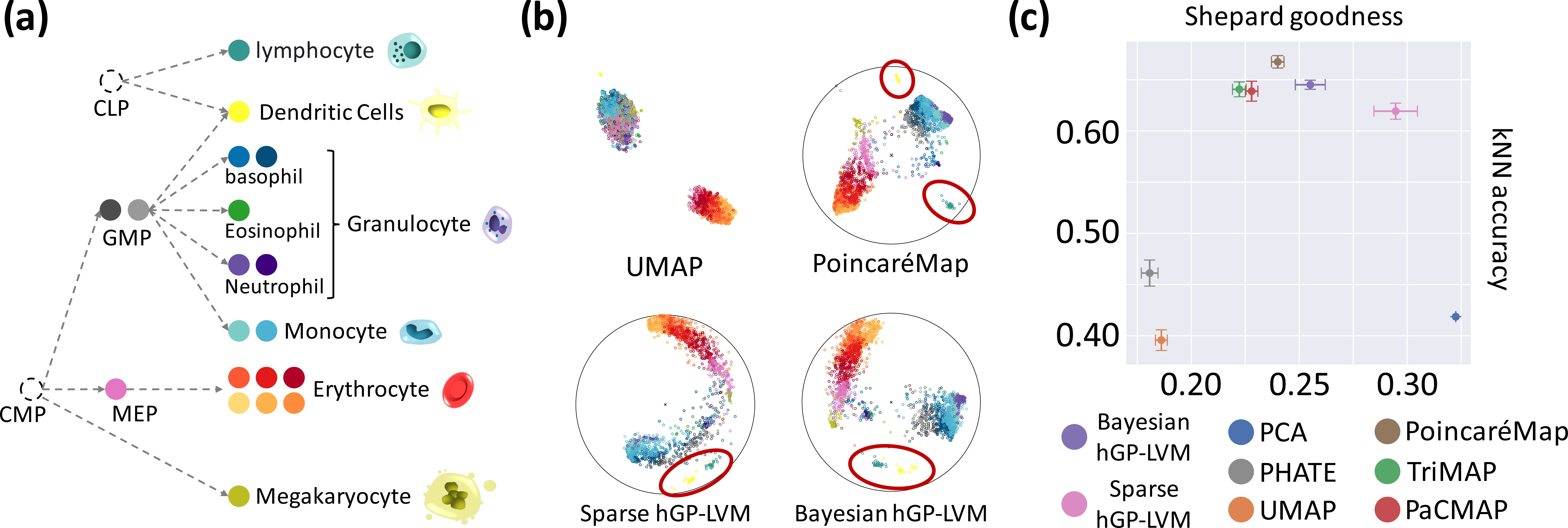}
    }
    \vspace{-0.1cm}
    \caption{\textbf{Experimental results on the scRNA-seq dataset.} (a) The canonical hematopoietic cell lineage tree. (b) Two-dimensional embedding of UMAP, Poincar\'{e}Map, Sparse hGP-LVM, and Bayesian hGP-LVM. The colors correspond to those of the lineage tree. (c) The error bar plot of the compared methods. The same experiment was conducted 30 times, and we computed the mean error with the standard deviation.}
    \label{fig:main_exp_paul}
\end{figure*}

\paragraph{Embedding Comparison with Different Length Scales}
As mentioned in Section~\ref{subsec:hyperboloid_kernel}, \textit{the meaning of the length scale $\kappa$ is how much we expect the latent variables to follow the hyperbolic curvature}. An experimental verification of this statement is shown in Figure~\ref{fig:main_bin_tree_result}~(b). Here, we confirm that the representation was spread and aligned as $\kappa$ increased. Although the embeddings of depths $1$ and $2$ were mixed around the origin for $\kappa=10$ and $\kappa=30$, they were separated for $\kappa=100$. The length scale parameter determined the range of the latent variables, and a large $\kappa$ value effectively brought the strong hyperbolic curvature to the latent representation. Thus, the $\kappa$ value must be determined according to the degree to which we expect the hierarchical structure in the data.

\subsection{Mouse Myeloid Progenitors Dataset}
\label{section:mouse_myeloid_progenitors}
We also conducted an experiment on the real-world scRNA-seq dataset presented in~\citep{paul2015transcriptional}. The synthesis data results are shown in Appendix A.5 to verify the effectiveness of the GP-based visualization approach. This scRNA-seq dataset exhibits multiple intermediate populations of cell populations ($N=2,730$) from the progenitors (CMP, GMP, MEP, and CLP) to the restricted myeloid cells, e.g., erythrocytes, leukocytes, and lymphocytes, representing the hierarchical structure along the cell differentiation. Figure~\ref{fig:main_exp_paul} (a) shows the canonical hematopoietic cell lineage tree. Each cell's features is a bag of gene expressions, and they have labels annotated with the clusters identified in the original study. In the current study, we preprocessed the original scRNA-seq data by selecting $1,000$ highly-variable genes ($D=1,000$) following the literature~\citep{klimovskaia2020poincare} and~\citep{zheng2017massively}. Note that the sample size is prohibitive for hGP-LVM, and we used sparse and Bayesian hGP-LVMs with $\kappa=100$ and $M=50$. We compared them with the UMAP and Poincar\'{e}Map methods qualitatively and further compared with PCA~\citep{hotelling1933analysis}, PHATE~\citep{moon2019visualizing}, TriMAP~\citep{amid2019trimap}, and PacMAP~\citep{wang2021understanding} quantitatively. We did not take GP-LVMs as compared methods since their learning did not work under the same conditions as hGP-LVM. In this evaluation, we employed the Shepard goodness to evaluate the quality of global preservation. However, the dataset was still noisy after preprocessing, and the local structure was not trustworthy; thus, to use trustworthy label information, we adopted the $k$-NN classification accuracy $(k=5)$ as the local metric.
\par
Figure~\ref{fig:main_exp_paul} (b) shows the visualization of the scRNA-seq dataset. As can be seen, UMAP's embedding was torn and mixed with the population of lymphocytes, dendric cells, granulocytes, and monocytes. In addition, the Poincar\'{e}Map embedding successfully separated the cell populations; however, the continuity starting from MEP and GMP was not evident, and there was a large distance between the population of lymphocytes and dendric cells, which are developed from the same progenitors, i.e., CLP. The proposed hGP-LVMs' embedding preserved the cell population and the continuity along the development, and the inter-population similarity of the CLP progenitors was preserved. In addition, we confirmed that the megakaryocyte population of the Bayesian hGP-LVM is more evident than the sparse hGP-LVM., which indicates that the uncertainty of the latent variables introduced by the full Bayesian inference is suitable for the noisy real-world dataset and facilitates effective embedding of the scRNA-seq dataset. Finally, the quantitative results are shown in Figure~\ref{fig:main_exp_paul} (c). Recall the tradeoff relationship between local and global preservation. Both preservations are required to embed the hierarchical data, and the proposed hGP-LVMs exhibit relatively high results for both metrics. 
\par
In summary, the results have confirmed the effectiveness of the proposed hGP-LVMs against hyperbolic generative models and neighbor embedding methods. Although point estimation methods work efficiently for synthetic data, Bayesian inference is required for noisy real-world datasets. Thus, in consideration of the tradeoff between time complexity and representation power, we must utilize such methods properly.

\section{Conclusion}
\label{section:Conclusion}
This paper has proposed the hyperbolic extension of GP-LVMs to realize the faithful low-dimensional embedding of the hierarchical data via nonparametric estimation. We established the HE kernel and incorporated the Riemannian optimization process with the previous sparse GP inference method. We have also introduced the reparameterization trick on the Lorentz model for a fully Bayesian estimation of the latent variables. The experiments conducted in this study validated the embedding accuracy of the proposed hGP-LVMs on synthetic hierarchical and real-world scRNA-seq datasets, and the results demonstrated the effectiveness of the GP-based modeling for visualization-aided DR.

\paragraph{Limitation and Future Work.}
We assume a hierarchical structure; however, if there is no clear hierarchical relation in a dataset, the proposed method cannot embed the observed data more effectively than general GP-LVM frameworks. In addition, the computational may be an issue because it is linear to the sample size and prohibitive for several huge datasets. The embedding quality of the neighbor embedding method scales with the sample size, and they are more effective for larger datasets.
\par
Therefore, future studies will focus on improving the Bayesian extension's time complexity and approximated inference. In addition, improving the initial embedding is also important in terms of realizing more efficient learning for the GP-based latent variable models.

\section*{Acknowledgement}
This study was supported in part by JSPS KAKENHI Grant Numbers JP24K02942, JP23K21676, JP23K11211, and JP24KJ0324.


\begin{thebibliography}{}

\bibitem[Amid and Warmuth, 2019]{amid2019trimap}
Amid, E. and Warmuth, M.~K. (2019).
\newblock {TriMap}: Large-scale dimensionality reduction using triplets.
\newblock {\em arXiv preprint arXiv:1910.00204}.

\bibitem[Atigh et~al., 2022]{atigh2022hyperbolic}
Atigh, M.~G., Schoep, J., Acar, E., Van~Noord, N., and Mettes, P. (2022).
\newblock Hyperbolic image segmentation.
\newblock In {\em Proceedings of the IEEE/CVF conference on Computer Vision and Pattern Recognition}, pages 4453--4462.

\bibitem[Azangulov et~al., 2022]{azangulov2022stationary}
Azangulov, I., Smolensky, A., Terenin, A., and Borovitskiy, V. (2022).
\newblock Stationary kernels and {Gaussian} processes on {Lie} groups and their homogeneous spaces i: the compact case.
\newblock {\em arXiv preprint arXiv:2208.14960}.

\bibitem[Azangulov et~al., 2023]{azangulov2023stationary}
Azangulov, I., Smolensky, A., Terenin, A., and Borovitskiy, V. (2023).
\newblock Stationary kernels and {Gaussian} processes on {Lie} groups and their homogeneous spaces ii: non-compact symmetric spaces.
\newblock {\em arXiv preprint arXiv:2301.13088}.

\bibitem[Bauer et~al., 2016]{bauer2016understanding}
Bauer, M., Van~der Wilk, M., and Rasmussen, C.~E. (2016).
\newblock Understanding probabilistic sparse {Gaussian} process approximations.
\newblock {\em Advances in Neural Information Processing Systems}, 29:1--9.

\bibitem[Becht et~al., 2019]{becht2019dimensionality}
Becht, E., McInnes, L., Healy, J., Dutertre, C.-A., Kwok, I.~W., Ng, L.~G., Ginhoux, F., and Newell, E.~W. (2019).
\newblock Dimensionality reduction for visualizing single-cell data using {UMAP}.
\newblock {\em Nature Biotechnology}, 37(1):38--44.

\bibitem[Borovitskiy et~al., 2020]{borovitskiy2020matern}
Borovitskiy, V., Terenin, A., Mostowsky, P., et~al. (2020).
\newblock Mat{\'e}rn {Gaussian} processes on {Riemannian} manifolds.
\newblock {\em Advances in Neural Information Processing Systems}, 33:12426--12437.

\bibitem[Chami et~al., 2020]{chami2020low}
Chami, I., Wolf, A., Juan, D.-C., Sala, F., Ravi, S., and R{\'e}, C. (2020).
\newblock Low-dimensional hyperbolic knowledge graph embeddings.
\newblock {\em arXiv preprint arXiv:2005.00545}.

\bibitem[Cho et~al., 2022]{cho2022rotated}
Cho, S., Lee, J., Park, J., and Kim, D. (2022).
\newblock A rotated hyperbolic wrapped normal distribution for hierarchical representation learning.
\newblock {\em Advances in Neural Information Processing Systems}, 35:17831--17843.

\bibitem[de~Souza et~al., 2021]{de2021learning}
de~Souza, D., Mesquita, D., Gomes, J.~P., and Mattos, C.~L. (2021).
\newblock Learning {GPLVM} with arbitrary kernels using the unscented transformation.
\newblock In {\em Proceedings of the International Conference on Artificial Intelligence and Statistics}, pages 451--459.

\bibitem[Desai et~al., 2023]{desai2023hyperbolic}
Desai, K., Nickel, M., Rajpurohit, T., Johnson, J., and Vedantam, S.~R. (2023).
\newblock Hyperbolic image-text representations.
\newblock In {\em Proceedings of the International Conference on Machine Learning}, pages 7694--7731.

\bibitem[Fan et~al., 2023]{fan2023horospherical}
Fan, X., Yang, C.-H., and Vemuri, B. (2023).
\newblock Horospherical decision boundaries for large margin classification in hyperbolic space.
\newblock {\em Advances in Neural Information Processing Systems}, 36:1--11.

\bibitem[Fang et~al., 2021]{fang2021kernel}
Fang, P., Harandi, M., and Petersson, L. (2021).
\newblock Kernel methods in hyperbolic spaces.
\newblock In {\em Proceedings of the IEEE/CVF International Conference on Computer Vision}, pages 10665--10674.

\bibitem[Feragen et~al., 2015]{feragen2015geodesic}
Feragen, A., Lauze, F., and Hauberg, S. (2015).
\newblock Geodesic exponential kernels: When curvature and linearity conflict.
\newblock In {\em Proceedings of the IEEE Conference on Computer Vision and Pattern Recognition}, pages 3032--3042.

\bibitem[Ganea et~al., 2018]{ganea2018hyperbolic}
Ganea, O., B{\'e}cigneul, G., and Hofmann, T. (2018).
\newblock Hyperbolic neural networks.
\newblock {\em Advances in Neural Information Processing Systems}, 31:1--11.

\bibitem[Higgins et~al., 2017]{higgins2017betavae}
Higgins, I., Matthey, L., Pal, A., Burgess, C., Glorot, X., Botvinick, M., Mohamed, S., and Lerchner, A. (2017).
\newblock beta-{VAE}: Learning basic visual concepts with a constrained variational framework.
\newblock In {\em Proceedings of the International Conference on Learning Representations}, pages 1--22.

\bibitem[Hotelling, 1933]{hotelling1933analysis}
Hotelling, H. (1933).
\newblock Analysis of a complex of statistical variables into principal components.
\newblock {\em Journal of Educational Psychology}, 24(6):417--441.

\bibitem[Istas, 2012]{istas2012manifold}
Istas, J. (2012).
\newblock Manifold indexed fractional fields*.
\newblock {\em ESAIM: Probability and Statistics}, 16:222--276.

\bibitem[Jaquier et~al., 2022]{jaquier2022bringing}
Jaquier, N., Rozo, L., Gonz{\'a}lez-Duque, M., Borovitskiy, V., and Asfour, T. (2022).
\newblock Bringing robotics taxonomies to continuous domains via {GPLVM} on hyperbolic manifolds.
\newblock {\em arXiv preprint arXiv:2210.01672}.

\bibitem[Jensen et~al., 2020]{jensen2020manifold}
Jensen, K., Kao, T.-C., Tripodi, M., and Hennequin, G. (2020).
\newblock Manifold {GPLVMs} for discovering non-{Euclidean} latent structure in neural data.
\newblock {\em Advances in Neural Information Processing Systems}, 33:22580--22592.

\bibitem[Kingma and Welling, 2013]{kingma2013auto}
Kingma, D.~P. and Welling, M. (2013).
\newblock Auto-encoding variational {Bayes}.
\newblock {\em arXiv preprint arXiv:1312.6114}.

\bibitem[Klimovskaia et~al., 2020]{klimovskaia2020poincare}
Klimovskaia, A., Lopez-Paz, D., Bottou, L., and Nickel, M. (2020).
\newblock Poincar{\'e} maps for analyzing complex hierarchies in single-cell data.
\newblock {\em Nature Communications}, 11(1):2966.

\bibitem[Kobak and Berens, 2019]{kobak2019art}
Kobak, D. and Berens, P. (2019).
\newblock The art of using {t-SNE} for single-cell transcriptomics.
\newblock {\em Nature Communications}, 10(1):5416.

\bibitem[Lalchand et~al., 2022]{lalchand2022generalised}
Lalchand, V., Ravuri, A., and Lawrence, N.~D. (2022).
\newblock Generalised {GPLVM} with stochastic variational inference.
\newblock In {\em Proceedings of the International Conference on Artificial Intelligence and Statistics}, pages 7841--7864.

\bibitem[Lawrence, 2005]{lawrence2005probabilistic}
Lawrence, N. (2005).
\newblock Probabilistic non-linear principal component analysis with {Gaussian} process latent variable models.
\newblock {\em Journal of Machine Learning Research}, 6(11):1--34.

\bibitem[Lawrence, 2007]{lawrence2007learning}
Lawrence, N.~D. (2007).
\newblock Learning for larger datasets with the {Gaussian} process latent variable model.
\newblock In {\em Proceedings of the International Conference on Artificial Intelligence and Statistics}, pages 243--250.

\bibitem[Luecken and Theis, 2019]{luecken2019current}
Luecken, M.~D. and Theis, F.~J. (2019).
\newblock Current best practices in single-cell {RNA}-seq analysis: a tutorial.
\newblock {\em Molecular Systems Biology}, 15(6):e8746.

\bibitem[Mallasto and Feragen, 2018]{mallasto2018wrapped}
Mallasto, A. and Feragen, A. (2018).
\newblock Wrapped {Gaussian} process regression on {Riemannian} manifolds.
\newblock In {\em Proceedings of the IEEE Conference on Computer Vision and Pattern Recognition}, pages 5580--5588.

\bibitem[Mathieu et~al., 2019]{mathieu2019continuous}
Mathieu, E., Le~Lan, C., Maddison, C.~J., Tomioka, R., and Teh, Y.~W. (2019).
\newblock Continuous hierarchical representations with {Poincar{\'e}} variational auto-encoders.
\newblock {\em Advances in Neural Information Processing Systems}, 32:1--12.

\bibitem[McInnes et~al., 2018]{mcinnes2018umap}
McInnes, L., Healy, J., and Melville, J. (2018).
\newblock {UMAP}: Uniform manifold approximation and projection for dimension reduction.
\newblock {\em arXiv preprint arXiv:1802.03426}.

\bibitem[Moon et~al., 2019]{moon2019visualizing}
Moon, K.~R., Van~Dijk, D., Wang, Z., Gigante, S., Burkhardt, D.~B., Chen, W.~S., Yim, K., Elzen, A. v.~d., Hirn, M.~J., Coifman, R.~R., et~al. (2019).
\newblock Visualizing structure and transitions in high-dimensional biological data.
\newblock {\em Nature Biotechnology}, 37(12):1482--1492.

\bibitem[Moreno-Mu{\~n}oz et~al., 2022]{moreno2022revisiting}
Moreno-Mu{\~n}oz, P., Feldager, C., and Hauberg, S. (2022).
\newblock Revisiting active sets for {Gaussian} process decoders.
\newblock {\em Advances in Neural Information Processing Systems}, 35:6603--6614.

\bibitem[Nagano et~al., 2019]{nagano2019wrapped}
Nagano, Y., Yamaguchi, S., Fujita, Y., and Koyama, M. (2019).
\newblock A wrapped normal distribution on hyperbolic space for gradient-based learning.
\newblock In {\em Proceedings of the International Conference on Machine Learning}, pages 4693--4702.

\bibitem[Nickel and Kiela, 2017]{nickel2017poincare}
Nickel, M. and Kiela, D. (2017).
\newblock Poincar{\'e} embeddings for learning hierarchical representations.
\newblock {\em Advances in Neural Information Processing Systems}, 30:1--10.

\bibitem[Nickel and Kiela, 2018]{nickel2018learning}
Nickel, M. and Kiela, D. (2018).
\newblock Learning continuous hierarchies in the {Lorentz} model of hyperbolic geometry.
\newblock In {\em Proceedings of the International Conference on Machine Learning}, pages 3779--3788.

\bibitem[Niu et~al., 2023]{niu2023intrinsic}
Niu, M., Dai, Z., Cheung, P., and Wang, Y. (2023).
\newblock Intrinsic {Gaussian} process on unknown manifolds with probabilistic metrics.
\newblock {\em Journal of Machine Learning Research}, 24(104):1--42.

\bibitem[Paul et~al., 2015]{paul2015transcriptional}
Paul, F., Arkin, Y., Giladi, A., Jaitin, D.~A., Kenigsberg, E., Keren-Shaul, H., Winter, D., Lara-Astiaso, D., Gury, M., Weiner, A., et~al. (2015).
\newblock Transcriptional heterogeneity and lineage commitment in myeloid progenitors.
\newblock {\em Cell}, 163(7):1663--1677.

\bibitem[Peng et~al., 2021]{peng2021hyperbolic}
Peng, W., Varanka, T., Mostafa, A., Shi, H., and Zhao, G. (2021).
\newblock Hyperbolic deep neural networks: A survey.
\newblock {\em IEEE Transactions on Pattern Analysis and Machine Intelligence}, 44(12):10023--10044.

\bibitem[Rasmussen and Williams, 2006]{rasmussen2006gaussian}
Rasmussen, C.~E. and Williams, C.~K. (2006).
\newblock {\em Gaussian Processes for Machine Learning}.
\newblock MIT Press.

\bibitem[Sala et~al., 2018]{sala2018representation}
Sala, F., De~Sa, C., Gu, A., and R{\'e}, C. (2018).
\newblock Representation tradeoffs for hyperbolic embeddings.
\newblock In {\em Proceedings of the International Conference on Machine Learning}, pages 4460--4469.

\bibitem[Salimbeni and Deisenroth, 2017]{salimbeni2017doubly}
Salimbeni, H. and Deisenroth, M. (2017).
\newblock Doubly stochastic variational inference for deep {Gaussian} processes.
\newblock {\em Advances in Neural Information Processing Systems}, 30:1--12.

\bibitem[Titsias, 2009]{titsias2009variational}
Titsias, M. (2009).
\newblock Variational learning of inducing variables in sparse {Gaussian} processes.
\newblock In {\em Proceedings of the International Conference on Artificial Intelligence and Statistics}, pages 567--574.

\bibitem[Titsias and Lawrence, 2010]{titsias2010bayesian}
Titsias, M. and Lawrence, N.~D. (2010).
\newblock Bayesian {Gaussian} process latent variable model.
\newblock In {\em Proceedings of the International Conference on Artificial Intelligence and Statistics}, pages 844--851.

\bibitem[Van~der Maaten and Hinton, 2008]{van2008visualizing}
Van~der Maaten, L. and Hinton, G. (2008).
\newblock Visualizing data using {t-SNE}.
\newblock {\em Journal of Machine Learning Research}, 9(11):1--25.

\bibitem[Wang et~al., 2021]{wang2021understanding}
Wang, Y., Huang, H., Rudin, C., and Shaposhnik, Y. (2021).
\newblock Understanding how dimension reduction tools work: an empirical approach to deciphering {t-SNE, UMAP, TriMAP, and PaCMAP} for data visualization.
\newblock {\em Journal of Machine Learning Research}, 22(201):1--73.

\bibitem[Yang et~al., 2023]{yang2023expanding}
Yang, M., Fang, P., and Xue, H. (2023).
\newblock Expanding the hyperbolic kernels: a curvature-aware isometric embedding view.
\newblock In {\em Proceedings of the International Joint Conference on Artificial Intelligence}, pages 4469--4477.

\bibitem[Zheng et~al., 2017]{zheng2017massively}
Zheng, G.~X., Terry, J.~M., Belgrader, P., Ryvkin, P., Bent, Z.~W., Wilson, R., Ziraldo, S.~B., Wheeler, T.~D., McDermott, G.~P., Zhu, J., et~al. (2017).
\newblock Massively parallel digital transcriptional profiling of single cells.
\newblock {\em Nature Communications}, 8(1):14049.

\end{thebibliography}

\begin{thebibliography}{}

\bibitem[Cho et~al., 2022]{cho2022rotated}
Cho, S., Lee, J., Park, J., and Kim, D. (2022).
\newblock A rotated hyperbolic wrapped normal distribution for hierarchical representation learning.
\newblock {\em Advances in Neural Information Processing Systems}, 35:17831--17843.

\bibitem[Espadoto et~al., 2019]{espadoto2019toward}
Espadoto, M., Martins, R.~M., Kerren, A., Hirata, N.~S., and Telea, A.~C. (2019).
\newblock Toward a quantitative survey of dimension reduction techniques.
\newblock {\em IEEE Transactions on Visualization and Computer Graphics}, 27(3):2153--2173.

\bibitem[{GPy}, 2012]{gpy}
{GPy} (2012).
\newblock {GPy}: A {Gaussian} process framework in {Python}.
\newblock Available: \url{http://github.com/SheffieldML/GPy}.

\bibitem[Joia et~al., 2011]{joia2011local}
Joia, P., Coimbra, D., Cuminato, J.~A., Paulovich, F.~V., and Nonato, L.~G. (2011).
\newblock Local affine multidimensional projection.
\newblock {\em IEEE Transactions on Visualization and Computer Graphics}, 17(12):2563--2571.

\bibitem[Klimovskaia et~al., 2020]{klimovskaia2020poincare}
Klimovskaia, A., Lopez-Paz, D., Bottou, L., and Nickel, M. (2020).
\newblock Poincar{\'e} maps for analyzing complex hierarchies in single-cell data.
\newblock {\em Nature Communications}, 11(1):2966.

\bibitem[McInnes et~al., 2018]{mcinnes2018umap}
McInnes, L., Healy, J., and Melville, J. (2018).
\newblock {UMAP}: Uniform manifold approximation and projection for dimension reduction.
\newblock {\em arXiv preprint arXiv:1802.03426}.

\bibitem[Sloane, 2007]{sloane2007line}
Sloane, N.~J. (2007).
\newblock The on-line encyclopedia of integer sequences.
\newblock In {\em Proceedings of the International Conference on Towards Mechanized Mathematical Assistants}, pages 130--130.

\bibitem[Venna and Kaski, 2001]{venna2001neighborhood}
Venna, J. and Kaski, S. (2001).
\newblock Neighborhood preservation in nonlinear projection methods: An experimental study.
\newblock In {\em Proceedings of the International Conference on Artificial Neural Networks}, pages 485--491.

\bibitem[Zu and Tao, 2022]{zu2022spacemap}
Zu, X. and Tao, Q. (2022).
\newblock {SpaceMAP}: Visualizing high-dimensional data by space expansion.
\newblock In {\em Proceedings of the International Conference on Machine Learning}, pages 27707--27723.

\end{thebibliography}

\section*{Checklist}

 \begin{enumerate}

 \item For all models and algorithms presented, check if you include:
 \begin{enumerate}
   \item A clear description of the mathematical setting, assumptions, algorithm, and/or model. [Yes] In Section~\ref{section:Backgrounds} and Appendix A.1.
   \item An analysis of the properties and complexity (time, space, sample size) of any algorithm. [Yes] In the last paragraph in Section~\ref{section:Methods}.
   \item (Optional) Anonymized source code, with specification of all dependencies, including external libraries. [Yes] In the supplemental material.
 \end{enumerate}

 \item For any theoretical claim, check if you include:
 \begin{enumerate}
   \item Statements of the full set of assumptions of all theoretical results. [Yes] In Section~\ref{section:Methods}.
   \item Complete proofs of all theoretical results. [Yes] In Appendix A.2.
   \item Clear explanations of any assumptions. [Yes] In Section~\ref{section:Backgrounds} and~\ref{section:Methods}.  
 \end{enumerate}

 \item For all figures and tables that present empirical results, check if you include:
 \begin{enumerate}
   \item The code, data, and instructions needed to reproduce the main experimental results (either in the supplemental material or as a URL). [Yes] In the supplemental material.
   \item All the training details (e.g., data splits, hyperparameters, how they were chosen). [Yes] In the first paragraphs in Section~\ref{section: binary_tree}, Section~\ref{section:mouse_myeloid_progenitors}, and Appendix A.4.
         \item A clear definition of the specific measure or statistics and error bars (e.g., with respect to the random seed after running experiments multiple times). [Yes] In the experimental results in Section~\ref{section:experiments} and Appendix A.5.
         \item A description of the computing infrastructure used. (e.g., type of GPUs, internal cluster, or cloud provider). [Yes] In Appendix A.4.
 \end{enumerate}

 \item If you are using existing assets (e.g., code, data, models) or curating/releasing new assets, check if you include:
 \begin{enumerate}
   \item Citations of the creator If your work uses existing assets. [Yes] In Appendix A.4.
   \item The license information of the assets, if applicable. [Yes] In Appendix A.4.
   \item New assets either in the supplemental material or as a URL, if applicable. [Yes] In the supplemental material.
   \item Information about consent from data providers/curators. [Yes] We use open-source datasets.
   \item Discussion of sensible content if applicable, e.g., personally identifiable information or offensive content. [Not Applicable]
 \end{enumerate}

 \item If you used crowdsourcing or conducted research with human subjects, check if you include:
 \begin{enumerate}
   \item The full text of instructions given to participants and screenshots. [Not Applicable]
   \item Descriptions of potential participant risks, with links to Institutional Review Board (IRB) approvals if applicable. [Not Applicable]
   \item The estimated hourly wage paid to participants and the total amount spent on participant compensation. [Not Applicable]
 \end{enumerate}

 \end{enumerate}


\renewcommand{\thesection}{A.\arabic{section}} 
\renewcommand{\theequation}{A.\arabic{equation}}
\renewcommand{\thefigure}{A.\arabic{figure}}
\setcounter{equation}{0}

\onecolumn
\aistatstitle{Hyperboloid GPLVM for Discovering\\ Continuous Hierarchies via Nonparametric Estimation:\\ Supplementary Materials}
\section{Derivation of Objectives of GP-LVMs}
\label{appendix:derivation_gplvm}
We first recall the GP-LVM model definition as
\begin{align}
    &\bm{\mathrm{y}}_{:,d}=\bm{\mathrm{f}}_{d}(\bm{\mathrm{X}})+\bm{\epsilon},\;\bm{\epsilon}\sim \mathcal{N}(\bm{0},\beta^{-1}\bm{\mathrm{I}}_{n}),\;\bm{\mathrm{f}}_{d}\sim\mathcal{GP}(\bm{0},k(\cdot,\, \cdot)).\label{eq_apx:gaussian_process}
\end{align}
The equal expression of Eq.~\eqref{eq_apx:gaussian_process} as the probabilistic density function is given by 
\begin{align*}
    p(\bm{\mathrm{y}}_{:,d}|\bm{\mathrm{f}}_{d})&=\mathcal{N}(\bm{\mathrm{y}}_{:,d}|\bm{\mathrm{f}}_{d},\,\beta^{-1}\bm{\mathrm{I}}_{n}),\\
    p(\bm{\mathrm{f}}_{d}|\bm{\mathrm{X}})&=\mathcal{N}(\bm{\mathrm{f}}_{d}|\bm{0},\,\bm{\mathrm{K}}_{nn}).
\end{align*}
Then, we derive the objective log-likelihood function of GP-LVM by marginalizing the GP prior $\bm{\mathrm{f}}_{d}$ as follows:
\begin{align}
    p(\bm{\mathrm{Y}}|\bm{\mathrm{X}})&=\prod_{d=1}^{D}\int p(\bm{\mathrm{y}}_{:,d}|\bm{\mathrm{f}}_{d})p(\bm{\mathrm{f}}_{d}|\bm{\mathrm{X}})d\bm{\mathrm{f}}_{d}\notag\\
    &=-\frac{ND}{2}\log2\pi-\frac{D}{2}\log |\bm{\mathrm{K}}_{nn}+\beta^{-1}\bm{\mathrm{I}}_{n}|-\frac{1}{2}\mathrm{tr}\left[(\bm{\mathrm{K}}_{nn}+\beta^{-1}\bm{\mathrm{I}}_{n})^{-1}\bm{\mathrm{Y}}\bm{\mathrm{Y}}^{\top}\right].\label{eq_apx:GP-LVM}
\end{align}
The practical problems of the GP-LVM objective in Eq.~\eqref{eq_apx:GP-LVM} are the computation of the $N \times N$ matrix inversion of $(\bm{\mathrm{K}}_{nn}+\beta^{-1}\bm{\mathrm{I}}_{n})$ and log-determinant $\log |\bm{\mathrm{K}}_{nn}+\beta^{-1}\bm{\mathrm{I}}_{n}|$, which are compressed into the Cholesky decomposition of $(\bm{\mathrm{K}}_{nn}+\beta^{-1}\bm{\mathrm{I}}_{n})$ with $O(\frac{1}{3}N^{3})$ time complexity. 
\par
\vfill
Next, we adopt the inducing method to derive the sparse GP-LVM objective. Formally, we assume the inducing points $\bm{\mathrm{u}}_{d}$, which are sufficient statistics of the prior $\bm{\mathrm{f}}_{d}$. Therefore, the joint distribution omitted $\bm{\mathrm{X}}$ and $\bm{\mathrm{Z}}$ are given by 
\begin{align*}
p(\bm{\mathrm{y}}_{:,d},\,\bm{\mathrm{f}}_{d},\,\bm{\mathrm{u}}_{d})=p(\bm{\mathrm{y}}_{:,d}|\bm{\mathrm{f}}_{d},\,\bm{\mathrm{u}}_{d})p(\bm{\mathrm{f}}_{d}|\bm{\mathrm{u}}_{d})p(\bm{\mathrm{u}}_{d})\approx p(\bm{\mathrm{y}}_{:,d}|\bm{\mathrm{u}}_{d})p(\bm{\mathrm{f}}_{d}|\bm{\mathrm{u}}_{d})p(\bm{\mathrm{u}}_{d}).
\end{align*}
The marginal log-likelihood of the sparse GP model is given by
\begin{align}
    p(\bm{\mathrm{Y}}|\bm{\mathrm{X}},\,\bm{\mathrm{Z}})=\prod_{d=1}^{D}\int p(\bm{\mathrm{y}}_{:,d}|\bm{\mathrm{u}}_{d},\,\bm{\mathrm{X}},\,\bm{\mathrm{Z}})p(\bm{\mathrm{u}}_{d}|\bm{\mathrm{Z}})d\bm{\mathrm{u}}_{d},\label{eq_apx:likelihood_sparse_GPLVM}
\end{align}
where
\begin{align*}
    p(\bm{\mathrm{y}}_{:,d}|\bm{\mathrm{f}}_{d})&=\mathcal{N}(\bm{\mathrm{y}}_{:,d}|\bm{\mathrm{f}}_{d},\,\beta^{-1}\bm{\mathrm{I}}_{n}),\\
    p(\bm{\mathrm{f}}_{d}|\bm{\mathrm{u}}_{d},\,\bm{\mathrm{X}},\,\bm{\mathrm{Z}})&=\mathcal{N}(\bm{\mathrm{f}}_{d}|\bm{\mathrm{K}}_{nm}\bm{\mathrm{K}}_{mm}^{-1}\bm{\mathrm{u}}_{d},\,\bm{\mathrm{K}}_{nn}-\bm{\mathrm{K}}_{nm}\bm{\mathrm{K}}_{mm}^{-1}\bm{\mathrm{K}}_{mn}),\\
    p(\bm{\mathrm{y}}_{:,d}|\bm{\mathrm{u}}_{d},\,\bm{\mathrm{X}},\,\bm{\mathrm{Z}})&=\mathcal{N}(\bm{\mathrm{y}}_{:,d}|\bm{\mathrm{K}}_{nm}\bm{\mathrm{K}}_{mm}^{-1}\bm{\mathrm{u}}_{d},\,\bm{\mathrm{K}}_{nn}-\bm{\mathrm{K}}_{nm}\bm{\mathrm{K}}_{mm}^{-1}\bm{\mathrm{K}}_{mn}+\beta^{-1}\bm{\mathrm{I}}_{n}),\\
    p(\bm{\mathrm{u}}_{d}|\bm{\mathrm{Z}})&=\mathcal{N}(\bm{\mathrm{f}}_{d}|\bm{0},\,\bm{\mathrm{K}}_{mm}).
\end{align*}
Then, we apply the variational method with Jensen's inequality and evaluate the lower bound of the log-likelihood function as
\begin{align*}
    \log p(\bm{\mathrm{y}}_{:,d}|\bm{\mathrm{u}}_{d},\,\bm{\mathrm{X}},\,\bm{\mathrm{Z}})&\geq\mathbb{E}_{p(\bm{\mathrm{f}}_{d}|\bm{\mathrm{u}}_{d},\,\bm{\mathrm{X}},\,\bm{\mathrm{Z}})}\left[\log p(\bm{\mathrm{y}}_{:,d}|\bm{\mathrm{f}}_{d})\right]\\
    &\triangleq\mathcal{F}_{1}.
\end{align*}
Thus, we obtain the following lower bound of the log-likelihood as
\begin{align}
    \log p(\bm{\mathrm{Y}}|\bm{\mathrm{X}},\,\bm{\mathrm{Z}})&=\log\prod_{d=1}^{D}\int\exp\left[\log p(\bm{\mathrm{y}}_{:,d}|\bm{\mathrm{u}}_{d},\,\bm{\mathrm{X}},\,\bm{\mathrm{Z}})\right]p(\bm{\mathrm{u}}_{:,d}|\bm{\mathrm{Z}})d\bm{\mathrm{u}}_{d}\notag\\
    &\geq\sum_{d=1}^{D}\log\mathcal{N}(\bm{\mathrm{y}}_{:,d}|\bm{0},\,\bm{\mathrm{K}}_{nm}\bm{\mathrm{K}}_{mm}^{-1}\bm{\mathrm{K}}_{mn}+\beta^{-1}\bm{\mathrm{I}}_{n})-\frac{D}{2}\mathrm{tr}\left(\bm{\mathrm{K}}_{nn}-\bm{\mathrm{K}}_{nm}\bm{\mathrm{K}}_{mm}^{-1}\bm{\mathrm{K}}_{mn}\right)\notag\\
    &=-\frac{D}{2}\log |\bm{\mathrm{Q}}_{nn}+\beta^{-1}\bm{\mathrm{I}}_{n}|-\frac{1}{2}\mathrm{tr}\left[(\bm{\mathrm{Q}}_{nn}+\beta^{-1}\bm{\mathrm{I}}_{n})^{-1}\bm{\mathrm{Y}}\bm{\mathrm{Y}}^{\top}\right]-\frac{\beta D}{2}\mathrm{tr}(\bm{\mathrm{K}}_{nn}-\bm{\mathrm{Q}}_{nn})\label{eq_apx:sparse_gplvm_before_extention},
\end{align}
where $\bm{\mathrm{Q}}_{nn}=\bm{\mathrm{K}}_{nm}\bm{\mathrm{K}}_{mm}^{-1}\bm{\mathrm{K}}_{mn}$. Here, we extend Eq.~\eqref{eq_apx:sparse_gplvm_before_extention} by applying the matrix determinant and inversion lemmas as
\begin{align}
    |\bm{\mathrm{K}}_{nm}\bm{\mathrm{K}}_{mm}^{-1}\bm{\mathrm{K}}_{mn}+\beta^{-1}\bm{\mathrm{I}}_{n}|&=|\bm{\mathrm{K}}_{mm}||\bm{\mathrm{K}}_{mm}+\beta\bm{\mathrm{K}}_{mn}\bm{\mathrm{K}}_{nm}|\label{eq_apx:det_lemma},\\
    (\bm{\mathrm{K}}_{nm}\bm{\mathrm{K}}_{mm}^{-1}\bm{\mathrm{K}}_{mn}+\beta^{-1}\bm{\mathrm{I}}_{n})^{-1}&=\beta\bm{\mathrm{I}}_{n}-\beta^{2}\bm{\mathrm{K}}_{nm}(\bm{\mathrm{K}}_{mm}+\beta\bm{\mathrm{K}}_{mn}\bm{\mathrm{K}}_{nm})^{-1}\bm{\mathrm{K}}_{mn}\label{eq_apx:inv_lemma}.
\end{align}
Then, we can obtain the sparse GP-LVM objectives by substituting Eqns.~\eqref{eq_apx:det_lemma} and \eqref{eq_apx:inv_lemma} into Eq.~\eqref{eq_apx:sparse_gplvm_before_extention} as
\begin{align*}
    \acute{\mathcal{F}}&=-\frac{D}{2}\log\frac{(2\pi)^{N}|\bm{\mathrm{A}}|}{\beta^{N}|\bm{\mathrm{K}}_{mm}|}-\frac{1}{2}\mathrm{tr}\left(\bm{\mathrm{W}}\bm{\mathrm{Y}}\bm{\mathrm{Y}}^{\top}\right)-\frac{\beta D}{2}\mathrm{tr}\left(\bm{\mathrm{K}}_{nn}\right)+\frac{\beta D}{2}\left(\bm{\mathrm{K}}_{mm}^{-1}\bm{\mathrm{K}}_{mn}\bm{\mathrm{K}}_{nm}\right),
\end{align*}
where $\bm{\mathrm{W}}=\beta\bm{\mathrm{I}}_{n}-\beta^{2}\bm{\mathrm{K}}_{nm}\bm{\mathrm{A}}^{-1}\bm{\mathrm{K}}_{mn}$ and $\bm{\mathrm{A}}=\bm{\mathrm{K}}_{mm}+\beta\bm{\mathrm{K}}_{mn}\bm{\mathrm{K}}_{nm}$.
\par
Finally, we derive the Bayesian GP-LVM objective. We first derive the lower bound straightforwardly as
\begin{align}
    &\log p(\bm{\mathrm{Y}}|\bm{\mathrm{Z}})\notag\\
    &=\sum_{d=1}^{D}\log\int\left\{\int p(\bm{\mathrm{y}}_{:,d}|\bm{\mathrm{u}}_{d},\,\bm{\mathrm{X}},\,\bm{\mathrm{Z}})p(\bm{\mathrm{u}}_{d})\mathrm{d}\bm{\mathrm{u}}_{d}\right\}p(\bm{\mathrm{X}})\mathrm{d}\bm{\mathrm{X}}\notag\\
    &=\sum_{d=1}^{D}\log\int\int\exp\left[\log\left\{p(\bm{\mathrm{y}}_{:,d}|\bm{\mathrm{u}}_{d},\,\bm{\mathrm{X}},\,\bm{\mathrm{Z}})\frac{p(\bm{\mathrm{X}})}{q(\bm{\mathrm{X}})}\right\}\right]q(\bm{\mathrm{X}})p(\bm{\mathrm{u}}_{d})d\bm{\mathrm{X}}\mathrm{d}\bm{\mathrm{u}}_{d}\notag\\
    &\geq\sum_{d=1}^{D}\log\int\exp\left[\mathbb{E}_{q(\bm{\mathrm{X}})}\left[\log p(\bm{\mathrm{y}}_{:.d}|\bm{\mathrm{u}}_{d},\,\bm{\mathrm{X}},\,\bm{\mathrm{Z}})\right]+\mathbb{E}_{q(\bm{\mathrm{X}})}\left[\log\frac{p(\bm{\mathrm{X}})}{q(\bm{\mathrm{X}})}\right]\right]p(\bm{\mathrm{u}}_{d})\mathrm{d}\bm{\mathrm{u}}_{d}\notag\\
    &=\sum_{d=1}^{D}\log\int\exp\left[\mathbb{E}_{q(\bm{\mathrm{X}})}\left[\log p(\bm{\mathrm{y}}_{:.d}|\bm{\mathrm{u}}_{d},\,\bm{\mathrm{X}},\,\bm{\mathrm{Z}})\right]\right]p(\bm{\mathrm{u}}_{d})d\bm{\mathrm{u}}_{d}-\mathrm{KL}[q(\bm{\mathrm{X}})||p(\bm{\mathrm{X}})]\notag\\
    &\geq\sum_{d=1}^{D}\log\int\exp\left[\mathbb{E}_{q(\bm{\mathrm{X}})}\left[\mathcal{F}_{1}\right]\right]p(\bm{\mathrm{u}}_{d})d\bm{\mathrm{u}}_{d}-\mathrm{KL}[q(\bm{\mathrm{X}})||p(\bm{\mathrm{X}})]\label{eq_apx:elbo_derivation_bgplvm}.
\end{align}
The expectation of $\mathcal{F}_{1}$ in Eq.~\eqref{eq_apx:elbo_derivation_bgplvm} can be computed in the closed form as
\begin{align}
    \mathbb{E}_{q(\bm{\mathrm{X}})}\left[\mathcal{F}_{1}\right]=-\frac{\beta}{2}\bm{\mathrm{y}}_{:,d}^{\top}\bm{\mathrm{y}}_{:,d}&+\beta\bm{\mathrm{y}}_{:,d}^{\top}\bm{\Psi}_{1}\bm{\mathrm{K}}_{mm}^{-1}\bm{\mathrm{u}}_{d}-\frac{\beta}{2}\bm{\mathrm{u}}_{d}^{\top}\bm{\mathrm{K}}_{mm}^{-1}\bm{\Psi}_{2}\bm{\mathrm{K}}_{mm}^{-1}\bm{\mathrm{u}}_{d}\notag\\
    &-\frac{\beta}{2}\psi_{0}+\frac{\beta}{2}\mathrm{tr}\left(\bm{\mathrm{K}}_{mm}^{-1}\bm{\Psi}_{2}\right)-\frac{N}{2}\log 2\pi\beta^{-1}.\label{eq_apx:E_q_F1}
\end{align}
By substituting Eq.~\eqref{eq_apx:E_q_F1} into the first term of Eq.~\eqref{eq_apx:elbo_derivation_bgplvm}, we obtain the following equation:
\begin{align}
        &\sum_{d=1}^{D}\log\int\exp\left[\mathbb{E}_{q(\bm{\mathrm{X}})}\left[\mathcal{F}_{1}\right]\right]p(\bm{\mathrm{u}}_{d})d\bm{\mathrm{u}}_{d}\notag\\
        &=\sum_{d=1}^{D}\log\left[\frac{\sqrt{\beta^{N}|\bm{\mathrm{K}}_{MM}|}}{\sqrt{(2\pi)^{N}|\beta\bm{\Psi}_{2}+\bm{\mathrm{K}}_{MM}|}}\exp\left\{-\frac{1}{2}\bm{\mathrm{y}}_{:,d}^{\top}\bm{\mathrm{W}}_{b}\bm{\mathrm{y}}_{:,d}-\frac{\beta}{2}\psi_{0}+\frac{\beta}{2}\mathrm{tr}\left(\bm{\mathrm{K}}_{mm}^{-1}\bm{\Psi}_{2}\right)\right\}\right],\label{eq_apx:elbo_first_term}
\end{align}
where $\psi_{0}=\mathrm{tr}\left[\bm{\mathrm{K}}_{nn}\right]$, $\bm{\mathrm{W}}_{b}=\beta\bm{\mathrm{I}}_{n}-\beta^{2}\bm{\Psi}_{1}^{\top}\bm{\mathrm{A}}_{b}\bm{\Psi}_{1}$ and $\bm{\mathrm{A}}_{b}^{-1}=\bm{\mathrm{K}}_{mm}+\beta\bm{\Psi}_{2}$. Finally, we expand Eq.~\eqref{eq_apx:elbo_derivation_bgplvm} as
\begin{align}
    \acute{\mathcal{F}}_{b}=&-\frac{D}{2}\log\frac{(2\pi)^{N}|\bm{\mathrm{A}}_{b}|}{\beta^{N}|\bm{\mathrm{K}}_{mm}|}-\frac{1}{2}\mathrm{tr}\left(\bm{\mathrm{W}}_{b}\bm{\mathrm{Y}}\bm{\mathrm{Y}}^{\top}\right)-\frac{\beta D}{2}\psi_{0}+\frac{\beta D}{2}\mathrm{tr}\left(\bm{\mathrm{K}}_{mm}^{-1}\bm{\Psi}_{2}\right)-\sum_{i=1}^{N}\mathrm{KL}_{i}.
\end{align}

\section{Proof of Lemma 1}
\label{appendix:proof}
Recall that the Lemma is $d_{\mathcal{L}^{Q}}(\bm{\mathrm{x}}_{1}',\,\bm{\mathrm{x}}_{2}')\approx d_{E}(\bm{\mathrm{x}}_{1},\bm{\mathrm{x}}_{2})+O(d_{E}(\bm{\mathrm{x}}_{1},\bm{\mathrm{x}}_{2})^{3})$ if $||\bm{\mathrm{x}}_{1}||_{2} \approx 0$ and $||\bm{\mathrm{x}}_{2}||_{2} \approx 0$.

\begin{proof}

We use the following expansion:
\begin{align*}
    \sqrt{1+z}&=1+\frac{1}{2}z-\frac{1}{8}z^{2}+\cdots,\\
    \cosh^{-1}(z)&=\sqrt{2(z-1)}\{1-\frac{1}{12}(z-1)+\frac{3}{160}(z-1)^{2}+\cdots\}.
\end{align*}
The first equation is the Taylor expansion of $\sqrt{1+z}$ around $z=0$, and the second equation is derived from the Puiseux expansion of $\frac{\cosh^{-1}(z)}{\sqrt{2(z-1)}}$ around $z=1$~\citep{sloane2007line}. We first expand the Lorentzian inner product by substituting the Taylor expansion as
\begin{align*}
    -\left<\bm{\mathrm{x}}_{1},\bm{\mathrm{x}}_{2}\right>_{\mathcal{L}^{Q}}&=\sqrt{1+||\bm{\mathrm{x}}_{1}||_{2}^{2}}\sqrt{1+||\bm{\mathrm{x}}_{2}||_{2}^{2}}-\bm{\mathrm{x}}_{1}^{\top}\bm{\mathrm{x}}_{2}\\
    &=\left(1+\frac{1}{2}||\bm{\mathrm{x}}_{1}||_{2}^{2}-\frac{1}{8}||\bm{\mathrm{x}}_{1}||_{2}^{4}+\cdots\right)\left(1+\frac{1}{2}||\bm{\mathrm{x}}_{2}||_{2}^{2}-\frac{1}{8}||\bm{\mathrm{x}}_{2}||_{4}^{2}+\cdots\right)-\bm{\mathrm{x}}_{1}^{\top}\bm{\mathrm{x}}_{2}\\
    &\approx1+\frac{1}{2}||\bm{\mathrm{x}}_{1}||_{2}^{2}+\frac{1}{2}||\bm{\mathrm{x}}_{2}||_{2}^{2}-\bm{\mathrm{x}}_{1}^{\top}\bm{\mathrm{x}}_{2}\\
    &=1+\frac{1}{2}||\bm{\mathrm{x}}_{1}-\bm{\mathrm{x}}_{2}||_{2}^{2}\\
    &=1+\frac{1}{2}d_{E}(\bm{\mathrm{x}}_{1},\bm{\mathrm{x}}_{2})^{2}.
\end{align*}
and then we obtain
\begin{align*}
    d_{\mathcal{L}^{Q}}(\bm{\mathrm{x}}_{1},\bm{\mathrm{x}}_{2})&=\cosh^{-1}(-\left<\bm{\mathrm{x}}_{1},\bm{\mathrm{x}}_{2}\right>_{\mathcal{L}^{Q}})\\
    &\approx\cosh^{-1}\left(1+\frac{1}{2}d_{E}(\bm{\mathrm{x}}_{1},\bm{\mathrm{x}}_{2})^{2}\right)\\
    &=d_{E}(\bm{\mathrm{x}}_{1},\bm{\mathrm{x}}_{2})\left\{1-\frac{1}{24}d_{E}(\bm{\mathrm{x}}_{1},\bm{\mathrm{x}}_{2})^{2}+\frac{3}{640}d_{E}(\bm{\mathrm{x}}_{1},\bm{\mathrm{x}}_{2})^{4}+\cdots\right\}\\
    &=d_{E}(\bm{\mathrm{x}}_{1},\bm{\mathrm{x}}_{2})+O\left(d_{E}(\bm{\mathrm{x}}_{1},\bm{\mathrm{x}}_{2})^{3}\right).
\end{align*}
\end{proof}
\section{Differentiation of hGP-LVMs}
\label{appendix:derivatives}
We differentiate objectives to realize the gradient-based optimization. We use the chain rules similar to the previous GP-LVM and derive the differentiation w.r.t. gram matrices. Recall the objectives of GP-LVM $\mathcal{F}$, Sparse GP-LVM $\acute{\mathcal{F}}$, and Bayesian GP-LVM $\acute{\mathcal{F}}_{b}$ as

\begin{align*}
    \mathcal{F}&=-\frac{ND}{2}\log2\pi-\frac{D}{2}\log |\bm{\mathrm{K}}_{nn}+\beta^{-1}\bm{\mathrm{I}}_{n}|-\frac{1}{2}\mathrm{tr}\left[(\bm{\mathrm{K}}_{nn}+\beta^{-1}\bm{\mathrm{I}}_{n})^{-1}\bm{\mathrm{Y}}\bm{\mathrm{Y}}^{\top}\right],\\
    \acute{\mathcal{F}}&=-\frac{D}{2}\log\frac{(2\pi)^{N}|\bm{\mathrm{A}}|}{\beta^{N}|\bm{\mathrm{K}}_{mm}|}-\frac{1}{2}\mathrm{tr}\left(\bm{\mathrm{W}}\bm{\mathrm{Y}}\bm{\mathrm{Y}}^{\top}\right)-\frac{\beta D}{2}\mathrm{tr}\left(\bm{\mathrm{K}}_{nn}\right)+\frac{\beta D}{2}\left(\bm{\mathrm{K}}_{mm}^{-1}\bm{\mathrm{K}}_{mn}\bm{\mathrm{K}}_{nm}\right),\\
    \acute{\mathcal{F}}_{b}&=-\frac{D}{2}\log\frac{(2\pi)^{N}|\bm{\mathrm{A}}_{b}|}{\beta^{N}|\bm{\mathrm{K}}_{mm}|}-\frac{1}{2}\mathrm{tr}\left(\bm{\mathrm{W}}_{b}\bm{\mathrm{Y}}\bm{\mathrm{Y}}^{\top}\right)-\frac{\beta D}{2}\mathrm{tr}\left(\bm{\mathrm{K}}_{nn}\right)+\frac{\beta D}{2}\mathrm{tr}\left(\bm{\mathrm{K}}_{mm}^{-1}\bm{\Psi}_{2}\right)-\sum_{i=1}^{N}\mathrm{KL}_{i}.
\end{align*}
where $\bm{\mathrm{A}}=\bm{\mathrm{K}}_{mm}+\beta\bm{\mathrm{K}}_{mn}\bm{\mathrm{K}}_{nm}$, $\bm{\mathrm{W}}=\beta\bm{\mathrm{I}}_{n}-\beta^{2}\bm{\mathrm{K}}_{nm}\bm{\mathrm{A}}^{-1}\bm{\mathrm{K}}_{mn}$, $\bm{\mathrm{A}}_{b}=\bm{\mathrm{K}}_{mm}+\beta\bm{\Psi}_{2}$, $\bm{\mathrm{W}}_{b}=\beta\bm{\mathrm{I}}_{n}-\beta^{2}\bm{\mathrm{\Psi}}_{1}^{\top}\bm{\mathrm{A}}_{b}^{-1}\bm{\mathrm{\Psi}}_{1}$, $\bm{\Psi}_{1}=\mathbb{E}_{q(\bm{\mathrm{X}})}[\bm{\mathrm{K}}_{mn}]$, $\bm{\Psi}_{2}=\mathbb{E}_{q(\bm{\mathrm{X}})}[\bm{\mathrm{K}}_{mn}\bm{\mathrm{K}}_{nm}]$, and $\mathrm{KL}_{i}=\mathrm{KL}[q(\bm{\mathrm{x}}_{i})||p(\bm{\mathrm{x}}_{i})]$.
\subsection{Differentiation w.r.t. Gram Matrices}
\paragraph{GP-LVM}
The gram matrix in GP-LVM in Eq.~\eqref{eq_apx:GP-LVM} is $\bm{\mathrm{K}}_{nn}$. The differentiation w.r.t. $\bm{\mathrm{K}}_{nn}$ is given as follows: 
\begin{align}
    \frac{\partial\mathcal{F}}{\partial\bm{\mathrm{K}}_{nn}}=-\frac{D}{2}(\bm{\mathrm{K}}_{nn}+\beta^{-1}\bm{\mathrm{I}}_{n})^{-1}+\frac{1}{2}(\bm{\mathrm{K}}_{nn}+\beta^{-1}\bm{\mathrm{I}}_{n})^{-1}\bm{\mathrm{Y}}\bm{\mathrm{Y}}^{\top}(\bm{\mathrm{K}}_{nn}+\beta^{-1}\bm{\mathrm{I}}_{n})^{-1}.\label{eq_apx:dF_dKnn}
\end{align}

\paragraph{Sparse GP-LVM}
The gram matrices in sparse GP-LVM are $\bm{\mathrm{K}}_{mn}$ and $\bm{\mathrm{K}}_{mm}$. The differentiation w.r.t. $\bm{\mathrm{K}}_{mn}$ and $\bm{\mathrm{K}}_{mm}$ is given as follows:
\begin{align}
    \frac{\partial\acute{\mathcal{F}}}{\partial\bm{\mathrm{K}}_{mn}}&=-\beta D \bm{\mathrm{A}}^{-1}\bm{\mathrm{K}}_{mn}+\beta^{2}\bm{\mathrm{A}}^{-1}\bm{\mathrm{K}}_{mn}\bm{\mathrm{Y}}\bm{\mathrm{Y}}^{\top}-\beta^{3}\bm{\mathrm{A}}^{-1}\bm{\mathrm{K}}_{mn}\bm{\mathrm{Y}}\bm{\mathrm{Y}}^{\top}\bm{\mathrm{K}}_{nm}\bm{\mathrm{A}}^{-1}\bm{\mathrm{K}}_{mn}+\beta D\bm{\mathrm{K}}_{mm}^{-1}\bm{\mathrm{K}}_{mn},\label{eq_apx:dF_dKmn}\\
    \frac{\partial\acute{\mathcal{F}}}{\partial\bm{\mathrm{K}}_{mm}}&=\frac{D}{2}\bm{\mathrm{K}}_{mm}^{-1}-\frac{D}{2}\bm{\mathrm{A}}^{-1}-\frac{\beta^{2}}{2}\bm{\mathrm{A}}^{-1}\bm{\mathrm{K}}_{mn}\bm{\mathrm{Y}}\bm{\mathrm{Y}}^{\top}\bm{\mathrm{K}}_{nm}\bm{\mathrm{A}}^{-1}-\frac{\beta D}{2}\bm{\mathrm{K}}_{mm}^{-1}\bm{\mathrm{K}}_{mn}\bm{\mathrm{K}}_{nm}\bm{\mathrm{K}}_{mm}^{-1}.\label{eq_apx:dF_dKmm (sparse)}
\end{align}

\paragraph{Bayesian GP-LVM}The gram matrices in Bayesian GP-LVM is $\bm{\Psi}_{1}$, $\bm{\Psi}_{2}$, and $\bm{\mathrm{K}}_{mm}$. The differentiation w.r.t. $\bm{\Psi}_{1}$, $\bm{\Psi}_{2}$, and $\bm{\mathrm{K}}_{mm}$ is given as follows:
\begin{align}
    \frac{\partial\acute{\mathcal{F}}_{b}}{\partial\bm{\Psi}_{1}}&=\beta^{2}\bm{\mathrm{A}}_{b}\bm{\Psi}_{1}\bm{\mathrm{Y}}\bm{\mathrm{Y}}^{\top},\label{eq_apx:dF_dKpsi1}\\
    \frac{\partial\acute{\mathcal{F}}_{b}}{\partial\bm{\Psi}_{2}}&=-\frac{\beta D}{2}\bm{\mathrm{A}}_{b}^{-1}-\frac{\beta^{3}}{2}\bm{\mathrm{A}}_{b}^{-1}\bm{\Psi}_{1}\bm{\mathrm{Y}}\bm{\mathrm{Y}}^{\top}\bm{\Psi}_{1}^{\top}\bm{\mathrm{A}}_{b}^{-1}+\frac{\beta D}{2}\bm{\mathrm{K}}_{mm}^{-1},\label{eq_apx:dF_dKpsi2}\\
    \frac{\partial\acute{\mathcal{F}}_{b}}{\partial\bm{\mathrm{K}}_{mm}}&=\frac{D}{2}\bm{\mathrm{K}}_{mm}^{-1}-\frac{D}{2}\bm{\mathrm{A}}_{b}^{-1}-\frac{\beta^{2}}{2}\bm{\mathrm{A}}_{b}^{-1}\bm{\Psi}_{1}\bm{\mathrm{Y}}\bm{\mathrm{Y}}^{\top}\bm{\Psi}_{1}^{\top}\bm{\mathrm{A}}_{b}^{-1}-\frac{\beta D}{2}\bm{\mathrm{K}}_{mm}^{-1}\bm{\Psi}_{2}\bm{\mathrm{K}}_{mm}^{-1}\label{eq_apx:dF_dmm (bayesian)}.
\end{align}
We can compute the gradient w.r.t. kernel parameters as
\begin{align}
    \frac{\partial\mathcal{F}}{\partial \sigma}&=\frac{1}{\sigma}\mathrm{tr}\left(\frac{\partial\mathcal{F}}{\partial\bm{\mathrm{K}}_{nn}}\right),\label{dF_dsigma(GPLVM)}\\
    \frac{\partial\acute{\mathcal{F}}}{\partial \sigma}&=\frac{1}{\sigma}\mathrm{tr}\left(\frac{\partial\acute{\mathcal{F}}}{\partial\bm{\mathrm{K}}_{mn}}+\frac{\partial\acute{\mathcal{F}}}{\partial\bm{\mathrm{K}}_{mm}}\right),\label{dF_dsigma(sparse_GPLVM)}\\
    \frac{\partial\acute{\mathcal{F}}_{b}}{\partial \sigma}&=\frac{1}{\sigma}\mathrm{tr}\left(\frac{\partial\acute{\mathcal{F}}_{b}}{\partial\bm{\Psi}_{1}}+2\frac{\partial\acute{\mathcal{F}}_{b}}{\partial\bm{\Psi}_{2}}\right).\label{dF_dsigma(Bayesian_GPLVM)}
\end{align}

\subsection{Differentiation w.r.t. Latent Variables}
Next, we derive the differentiation w.r.t. latent variables. In GP-LVM and sparse GP-LVM, the differentiation can be derived by using the chain rule as 
\begin{align}
    \frac{\partial \mathcal{F}}{\partial x_{iq}}=\mathrm{tr}\left(\frac{\partial\mathcal{F}}{\partial\bm{\mathrm{K}}_{nn}}\frac{\partial\bm{\mathrm{K}}_{nn}}{\partial x_{iq}}\right),\;\frac{\partial \acute{\mathcal{F}}}{\partial x_{iq}}=\mathrm{tr}\left(\frac{\partial\mathcal{F}}{\partial\bm{\mathrm{K}}_{mn}}\frac{\partial\bm{\mathrm{K}}_{mn}}{\partial x_{iq}}\right).\label{eq_apx:df_dx}
\end{align}
We derive the differentiation of the gram matrix w.r.t. latent variables by entry-wise computation. The differentiation of $[\bm{\mathrm{K}}_{nn}]_{ij}=k(\bm{\mathrm{x}}_{i},\bm{\mathrm{x}}_{j})$ and $[\bm{\mathrm{K}}_{mn}]_{kj}=k(\bm{\mathrm{z}}_{k},\bm{\mathrm{x}}_{j})$ w.r.t. latent variables is given as follows:
\begin{align}
    \frac{\partial k_{\mathcal{L}^{Q}}(\bm{\mathrm{x}}_{i},\bm{\mathrm{x}}_{j})}{\partial x_{iq}}&=\frac{k_{\mathcal{L}^{Q}}(\bm{\mathrm{x}}_{i},\bm{\mathrm{x}}_{j})}{\kappa\sqrt{\left<\bm{\mathrm{x}}_{i},\bm{\mathrm{x}}_{j}\right>_{\mathcal{L}^{Q}}-1}}\left(x_{jq}-\frac{x_{j0}}{x_{i0}}x_{iq}\right),\label{eq_apx:dkij_dx}\\
    \frac{\partial k_{\mathcal{L}^{Q}}(\bm{\mathrm{z}}_{k},\bm{\mathrm{x}}_{j})}{\partial x_{iq}}&=\frac{k_{\mathcal{L}^{Q}}(\bm{\mathrm{z}}_{k},\bm{\mathrm{x}}_{j})}{\kappa\sqrt{\left<\bm{\mathrm{z}}_{k},\bm{\mathrm{x}}_{j}\right>_{\mathcal{L}^{Q}}-1}}\left(x_{jq}-\frac{x_{j0}}{z_{k0}}z_{kq}\right).\label{eq_apx:dkkj_dx}
\end{align}
\par
However, the differentiation of the Bayesian hGP-LVM objective is challenging due to the reparameterization as 
\begin{align*}
    \bm{\mathrm{x}}_{i}^{(h)}&=\mathrm{Exp}_{\bm{\mu}_{i}}(\bm{\mathrm{u}}_{i}^{(h)}),\;\bm{\mathrm{u}}_{i}^{(h)}=\mathrm{PT}_{\bm{\mu}_{0}\rightarrow\bm{\mu}_{i}}(\bm{\mathrm{v}}_{i}^{(h)}),\;\tilde{\bm{\mathrm{v}}}_{i}^{(h)}=\bm{\mathrm{S}}_{i}^{\frac{1}{2}}\bm{\zeta}^{(h)}_{i}.
\end{align*}
We first differentiate $\acute{\mathcal{F}}_{b}$ without the KL term. In our implementation, we use $\frac{\partial\acute{\mathcal{F}}_{b}}{\partial\bm{\Psi}_{1}}$ and $\frac{\partial\acute{\mathcal{F}}_{b}}{\partial\bm{\Psi}_{2}}$ and recall $\bm{\Psi}$ statistics as
\begin{align*}
    \bm{\Psi}_{1}=\sum_{h=1}^{H}\bm{\Psi}_{1}^{(h)},\;\;&\;[\bm{\Psi}_{1}^{(h)}]_{kj}=k(\bm{\mathrm{z}}_{k},\bm{\mathrm{x}}_{j}^{(h)}),\\
    \bm{\Psi}_{2}=\sum_{h=1}^{H}\bm{\Psi}_{2}^{(h)},\;\;&\;\bm{\Psi}_{2}^{(h)}=\sum_{i=1}^{N}\bm{\Psi}_{2}^{(h,i)},\;\;\;[\bm{\Psi}_{2}^{(h,i)}]_{kl}=k(\bm{\mathrm{z}}_{k},\bm{\mathrm{x}}_{i}^{(h)})k(\bm{\mathrm{z}}_{l},\bm{\mathrm{x}}_{i}^{(h)}).\\    
\end{align*}
The chain rules were used to compute the differentiation w.r.t. variational parameters as
\begin{align}
    \frac{\partial\acute{\mathcal{F}}_{b}}{\partial\mu_{iq}}&=\sum_{h=1}^{H}\mathrm{tr}\left(\frac{\partial\acute{\mathcal{F}}_{b}}{\partial\bm{\Psi}_{1}^{(h)}}\frac{\partial\bm{\Psi}_{1}^{(h)}}{\partial\mu_{iq}}+\frac{\partial\acute{\mathcal{F}}_{b}}{\partial\bm{\Psi}_{2}^{(h)}}\frac{\partial\bm{\Psi}_{2}^{(h)}}{\partial\mu_{iq}}\right),\label{eq_apx:dF_dmu}\\
    \frac{\partial\acute{\mathcal{F}}_{b}}{\partial s_{iq}}&=\sum_{h=1}^{H}\mathrm{tr}\left(\frac{\partial\acute{\mathcal{F}}_{b}}{\partial\bm{\Psi}_{1}^{(h)}}\frac{\partial\bm{\Psi}_{1}^{(h)}}{\partial s_{iq}}+\frac{\partial\acute{\mathcal{F}}_{b}}{\partial\bm{\Psi}_{2}^{(h)}}\frac{\partial\bm{\Psi}_{2}^{(h)}}{\partial s_{iq}}\right)\label{eq_apx:dF_ds}.
\end{align}
We chain the differentiation to the variational parameters via latent representation $\bm{\mathrm{x}}_{i}^{(h)}$ as

\begin{align}
    \frac{\partial[\bm{\Psi}_{1}^{(h)}]_{ki}}{\partial \mu_{iq}}&=\left(\frac{\partial[\bm{\Psi}_{1}^{(h)}]_{ki}}{\partial \bm{\mathrm{x}}_{i}^{(h)}}\right)^{\top}\frac{\partial \bm{\mathrm{x}}_{i}^{(h)}}{\partial \mu_{iq}},&
    \frac{\partial[\bm{\Psi}_{2}^{(h)}]_{kl}}{\partial\mu_{iq}}&=\left(\frac{\partial[\bm{\Psi}_{2}^{(h)}]_{kl}^{(h,i)}}{\partial \bm{\mathrm{x}}_{i}^{(h)}}\right)^{\top}\frac{\partial \bm{\mathrm{x}}_{i}^{(h)}}{\partial \mu_{iq}},\notag\\
    \frac{\partial[\bm{\Psi}_{1}^{(h)}]_{ki}}{\partial s_{iq}}&=\left(\frac{\partial[\bm{\Psi}_{1}^{(h)}]_{ki}}{\partial \bm{\mathrm{x}}_{i}^{(h)}}\right)^{\top}\frac{\partial \bm{\mathrm{x}}_{i}^{(h)}}{\partial s_{iq}},&  \frac{\partial[\bm{\Psi}_{2}^{(h)}]_{kl}}{\partial s_{iq}}&=\left(\frac{\partial[\bm{\Psi}_{2}^{(h)}]_{kl}^{(h,i)}}{\partial \bm{\mathrm{x}}_{i}^{(h)}}\right)^{\top}\frac{\partial \bm{\mathrm{x}}_{i}^{(h)}}{\partial s_{iq}},\notag
\end{align}
where the differentiation of the gram matrices is given as 
\begin{align}
    \frac{\partial[\bm{\Psi}_{1}^{(h)}]_{ki}}{\partial \bm{\mathrm{x}}_{i}^{(h)}}&=\frac{k_{\mathcal{L}^{Q}}(\bm{\mathrm{z}}_{k},\bm{\mathrm{x}}_{i}^{(h)})}{\kappa\sqrt{\left<\bm{\mathrm{z}}_{k},\bm{\mathrm{x}}_{i}^{(h)}\right>_{\mathcal{L}^{Q}}-1}}\left(x_{iq}^{(h)}-\frac{x_{i0}^{(h)}}{z_{k0}}z_{kq}\right),\label{eq_apx:dpsi1_dx}\\
    \frac{\partial[\bm{\Psi}_{2}^{(h,i)}]_{kl}}{\partial \bm{\mathrm{x}}_{i}^{(h)}}&=\frac{k_{\mathcal{L}^{Q}}(\bm{\mathrm{z}}_{k},\bm{\mathrm{x}}_{i}^{(h)})k_{\mathcal{L}^{Q}}(\bm{\mathrm{z}}_{l},\bm{\mathrm{x}}_{i}^{(h)})}{\kappa\sqrt{\left<\bm{\mathrm{z}}_{k},\bm{\mathrm{x}}_{i}^{(h)}\right>_{\mathcal{L}^{Q}}-1}}\left(x_{jq}^{(h)}-\frac{x_{j0}^{(h)}}{z_{k0}}z_{kq}\right)\notag\\
    &\hspace{2.5cm}+\frac{k_{\mathcal{L}^{Q}}(\bm{\mathrm{z}}_{k},\bm{\mathrm{x}}_{i}^{(h)})k_{\mathcal{L}^{Q}}(\bm{\mathrm{z}}_{l},\bm{\mathrm{x}}_{i}^{(h)})}{\kappa\sqrt{\left<\bm{\mathrm{z}}_{l},\bm{\mathrm{x}}_{i}^{(h)}\right>_{\mathcal{L}^{Q}}-1}}\left(x_{jq}^{(h)}-\frac{x_{j0}^{(h)}}{z_{l0}}z_{lq}\right)\label{eq_apx:dpsi2_dx}.
\end{align}
Finally, we derive the differentiation of $\bm{\mathrm{x}}_{i}^{(h)}$ w.r.t. variational parameters by chaining the reparameterization. Let $\mathbbl{1}_{q}\in\mathbb{R}^{Q+1}$ be a one-hot vector whose $(q+1)$-th element is $1$. We first derive w.r.t. variational mean $\mu_{iq}$ as
\begin{align}
    \frac{\partial \bm{\mathrm{x}}_{i}^{(h)}}{\partial \mu_{iq}}=&\cosh(||\bm{\mathrm{u}}_{i}^{(h)}||_{\mathcal{L}^{Q}})\mathbbl{1}_{q}+\frac{\partial}{\partial \mu_{iq}}\left\{\cosh(||\bm{\mathrm{u}}_{i}^{(h)}||_{\mathcal{L}^{Q}})\right\}\bm{\mu}_{i}\notag\\
    &+\frac{\partial}{\partial \mu_{iq}}\left\{\frac{\sinh(||\bm{\mathrm{u}}_{i}^{(h)}||_{\mathcal{L}^{Q}})}{||\bm{\mathrm{u}}_{i}^{(h)}||_{\mathcal{L}^{Q}}}\right\}\bm{\mathrm{u}}_{i}^{(h)}+\frac{\sinh(||\bm{\mathrm{u}}_{i}^{(h)}||_{\mathcal{L}^{Q}})}{||\bm{\mathrm{u}}_{i}^{(h)}||_{\mathcal{L}^{Q}}}\frac{\partial \bm{\mathrm{u}}_{i}^{(h)}}{\partial \mu_{iq}}.\label{eq_apx:dx_dmu}
\end{align}
We then compute the differentiation of the second and third terms in Eq.~\eqref{eq_apx:dx_dmu} as
\begin{align}
    \frac{\partial}{\partial \mu_{iq}}\left\{\cosh(||\bm{\mathrm{u}}_{i}^{(h)}||_{\mathcal{L}^{Q}})\right\}&=\frac{\sinh(||\bm{\mathrm{u}}_{i}^{(h)}||_{\mathcal{L}^{Q}})}{||\bm{\mathrm{u}}_{i}^{(h)}||_{\mathcal{L}^{Q}}}\cdot\hat{\bm{\mathrm{u}}}^{(h)\top}_{i}\frac{\partial \bm{\mathrm{u}}^{(h)}_{i}}{\partial \mu_{iq}},\label{eq_apx:dcosh_dmu}\\
    \frac{\partial}{\partial \mu_{iq}}\left\{\frac{\sinh(||\bm{\mathrm{u}}^{(h)}_{i}||_{\mathcal{L}^{Q}})}{||\bm{\mathrm{u}}^{(h)}_{i}||_{\mathcal{L}^{Q}}}\right\}&=\frac{\cosh(||\bm{\mathrm{u}}^{(h)}_{i}||_{\mathcal{L}^{Q}})||\bm{\mathrm{u}}^{(h)}_{i}||_{\mathcal{L}^{Q}}-\sinh(||\bm{\mathrm{u}}^{(h)}_{i}||_{\mathcal{L}^{Q}})}{||\bm{\mathrm{u}}^{(h)}_{i}||_{\mathcal{L}^{Q}}^{3}}\cdot\hat{\bm{\mathrm{u}}}_{i}^{(h)\top}\frac{\partial \bm{\mathrm{u}}_{i}^{(h)}}{\partial \mu_{iq}}\label{eq_apx:dsinh_dmu},
\end{align}
where $\hat{\bm{\mathrm{u}}}_{i}=[-u_{i0},\,u_{i1},\ldots,u_{iQ}]^{\top}$. The differentiation of $\bm{\mathrm{u}}_{i}^{(h)}$ given as
\begin{align}
    \frac{\partial \bm{\mathrm{u}}_{i}^{(h)}}{\partial \mu_{iq}}=
    \begin{cases}
        -\frac{\tilde{\bm{\mu}}_{i}^{\top}\tilde{\bm{\mathrm{v}}}_{i}^{(h)}}{(\mu_{i0}+1)^{2}}(\bm{\mu}_{0}+\bm{\mu}_{i})+\frac{\tilde{\bm{\mu}}_{i}^{\top}\tilde{\bm{\mathrm{v}}}_{i}^{(h)}}{\mu_{i0}+1}\mathbbl{1}_{0}\,&(q=0),\\
        \frac{\tilde{v}_{iq}}{\mu_{i0}+1}(\bm{\mu}_{0}+\bm{\mu}_{i})+\frac{\tilde{\bm{\mu}}_{i}^{\top}\tilde{\bm{\mathrm{v}}}^{(h)}_{i}}{\mu_{i0}+1}\mathbbl{1}_{q}\,&(q\neq 0).\
    \end{cases}\label{eq_apx:du_dmu}
\end{align}
Next, we derive the differentiation w.r.t. variational variance $s_{iq}$ as
\begin{align}
    \frac{\partial \bm{\mathrm{x}}_{i}}{\partial s_{iq}}&=\frac{\partial}{\partial s_{iq}}\left\{\cosh(||\bm{\mathrm{u}}_{i}^{(h)}||_{\mathcal{L}^{Q}})\right\}\bm{\mu}_{i}+\frac{\partial}{\partial s_{iq}}\left\{\frac{\sinh(||\bm{\mathrm{u}}_{i}^{(h)}||_{\mathcal{L}^{Q}})}{||\bm{\mathrm{u}}_{i}^{(h)}||_{\mathcal{L}^{Q}}}\right\}\bm{\mathrm{u}}_{i}^{(h)}+\frac{\sinh(||\bm{\mathrm{u}}_{i}^{(h)}||_{\mathcal{L}^{Q}})}{||\bm{\mathrm{u}}_{i}^{(h)}||_{\mathcal{L}^{Q}}}\frac{\partial \bm{\mathrm{u}}_{i}^{(h)}}{\partial s_{iq}}.\label{eq_apx:dx_ds}
\end{align}
We compute the first and second terms in Eq.~\eqref{eq_apx:dx_ds} as
\begin{align}
    \frac{\partial}{\partial s_{iq}}\left\{\cosh(||\bm{\mathrm{u}}_{i}^{(h)}||_{\mathcal{L}^{Q}})\right\}&=\frac{\sinh(||\bm{\mathrm{u}}_{i}^{(h)}||_{\mathcal{L}^{Q}})}{||\bm{\mathrm{u}}_{i}^{(h)}||_{\mathcal{L}^{Q}}}\cdot\hat{\bm{\mathrm{u}}}_{i}^{(h)\top}\frac{\partial \bm{\mathrm{u}}_{i}^{(h)}}{\partial s_{iq}},\label{eq_apx:dcosh_ds}\\
    \frac{\partial}{\partial s_{iq}}\left\{\frac{\sinh(||\bm{\mathrm{u}}_{i}^{(h)}||_{\mathcal{L}^{Q}})}{||\bm{\mathrm{u}}_{iq}^{(h)}||_{\mathcal{L}^{Q}}}\right\}&=\frac{\cosh(||\bm{\mathrm{u}}_{i}^{(h)}||_{\mathcal{L}^{Q}})||\bm{\mathrm{u}}_{i}^{(h)}||_{\mathcal{L}^{Q}}-\sinh(||\bm{\mathrm{u}}_{i}^{(h)}||_{\mathcal{L}^{Q}})}{||\bm{\mathrm{u}}_{i}^{(h)}||_{\mathcal{L}^{Q}}^{3}}\cdot\hat{\bm{\mathrm{u}}}_{i}^{(h)\top}\frac{\partial \bm{\mathrm{u}}_{i}^{(h)}}{\partial s_{iq}}\label{eq_apx:dsinh_ds}.
\end{align}
Then, the differentiation of $\bm{\mathrm{u}}_{i}$ w.r.t. $s_{iq}$ is derived as
\begin{align}
    \frac{\partial \bm{\mathrm{u}}_{i}^{(h)}}{\partial s_{iq}}=\zeta_{iq}^{(h)}\mathbbl{1}_{q}+\frac{\mu_{iq}\zeta_{iq}^{(h)}}{\mu_{i0}+1}(\bm{\mu}_{0}+\bm{\mu}_{i}).\label{eq_apx:du_ds}
\end{align}
Finally, we compute the differentiation of the KL term. We first extend the KL term with Monte Carlo approximations as
\begin{align}
    \mathrm{KL}_{i}&=\sum_{h=1}^{H}\left\{\log \mathcal{N}_{\mathcal{L}^{Q}}^{w}(\bm{\mathrm{x}}_{i}^{(h)}|\bm{\mu}_{i},\bm{\mathrm{S}}_{i})-\log \mathcal{N}_{\mathcal{L}^{Q}}^{w}(\bm{\mathrm{x}}_{i}^{(h)}|\bm{0},\bm{\mathrm{I}}_{n})\right\}\notag\\
    &=\sum_{h=1}^{H}\left\{\log \mathcal{N}(\bm{\mathrm{v}}_{i}^{(h)}|\bm{0},\bm{\mathrm{S}}_{i})-(Q-1)\log\frac{\sinh(||\bm{\mathrm{u}}_{i}^{(h)}||_{\mathcal{L}^{Q}})}{||\bm{\mathrm{u}}_{i}^{(h)}||_{\mathcal{L}^{Q}}}\right.\notag\\
    &\hspace{2.0cm}\left.-\log\mathcal{N}(\bm{\mathrm{v}}_{i}^{(h)}|\bm{0},\bm{\mathrm{I}}_{n})+(Q-1)\log\frac{\sinh(||\bm{\mathrm{v}}_{i}^{(h)}||_{\mathcal{L}^{Q}})}{||\bm{\mathrm{v}}_{i}^{^{(h)}}||_{\mathcal{L}^{Q}}}\right\},\label{eq_apx:KL_monte_carlo_with_wrapped_Gaussian}
\end{align}
where
\begin{align*}
    \log\mathcal{N}(\bm{\mathrm{v}}_{i}^{(h)}|\bm{0},\bm{\mathrm{S}}_{i})&=-\frac{Q}{2}\log 2\pi-\frac{1}{2}\log|\bm{\mathrm{S}}_{i}|-\frac{1}{2}\bm{\zeta}_{i}^{(h)\top}\bm{\zeta}_{i}^{(h)},\\
    \log\mathcal{N}(\bm{\mathrm{v}}_{i}^{(h)}|\bm{0},\bm{\mathrm{I}}_{n})&=-\frac{Q}{2}\log 2\pi-\frac{1}{2}\bm{\zeta}_{i}^{(h)\top}\bm{\mathrm{S}}_{i}\bm{\zeta}_{i}^{(h)}.
\end{align*}
Its differentiation can be computed using Eqns.~\eqref{eq_apx:du_dmu} and~\eqref{eq_apx:du_ds}. The differentiation of the second and fourth terms is given as
\begin{align*}
    \frac{\partial}{\partial \mu_{iq}}\left\{\log\frac{\sinh(||\bm{\mathrm{u}}_{i}^{(h)}||_{\mathcal{L}^{Q}})}{||\bm{\mathrm{u}}_{i}^{(h)}||_{\mathcal{L}^{Q}}}\right\}&=\left\{\frac{1}{\tanh(||\bm{\mathrm{u}}_{i}^{(h)}||_{\mathcal{L}^{Q}})||\bm{\mathrm{u}}_{i}^{(h)}||_{\mathcal{L}^{Q}}}-\frac{1}{||\bm{\mathrm{u}}_{i}^{(h)}||_{\mathcal{L}^{Q}}}\right\}\cdot\hat{\bm{\mathrm{u}}}_{i}^{(h)\top}\frac{\partial \bm{\mathrm{u}}_{i}^{(h)}}{\partial \mu_{iq}},\\
    \frac{\partial}{\partial s_{iq}}\left\{\log\frac{\sinh(||\bm{\mathrm{u}}_{i}^{(h)}||_{\mathcal{L}^{Q}})}{||\bm{\mathrm{u}}^{(h)}_{i}||_{\mathcal{L}^{Q}}}\right\}&=\left\{\frac{1}{\tanh(||\bm{\mathrm{u}}_{i}^{(h)}||_{\mathcal{L}^{Q}})||\bm{\mathrm{u}}_{i}^{(h)}||_{\mathcal{L}^{Q}}}-\frac{1}{||\bm{\mathrm{u}}_{i}^{(h)}||_{\mathcal{L}^{Q}}}\right\}\cdot\hat{\bm{\mathrm{u}}}_{i}^{(h)\top}\frac{\partial \bm{\mathrm{u}}_{i}^{(h)}}{\partial s_{iq}},\\
    \frac{\partial}{\partial s_{iq}}\left\{\log\frac{\sinh(||\bm{\mathrm{v}}_{i}^{(h)}||_{\mathcal{L}^{Q}})}{||\bm{\mathrm{v}}_{i}^{(h)}||_{\mathcal{L}^{Q}}}\right\}&=\left\{\frac{1}{\tanh(||\bm{\mathrm{u}}_{i}^{(h)}||_{\mathcal{L}^{Q}})||\bm{\mathrm{u}}_{i}^{(h)}||_{\mathcal{L}^{Q}}}-\frac{1}{||\bm{\mathrm{u}}_{i}^{(h)}||_{\mathcal{L}^{Q}}}\right\}\frac{\zeta_{iq}^{(h)}}{\sqrt{s_{iq}}}.
\end{align*}

\section{Experimental Details}
\label{appendix:experimental_details}
We implemented our methods partially using the GPy~\citep{gpy} source code (BSD-3-Clause license) and experimented partially using Poincar\'{e}Map (CC BY-NC 4.0 license)~\citep{klimovskaia2020poincare} and RotHVAE~\citep{cho2022rotated}, sincerely appreciating their contribution. Our code ran on a single Intel Core i7-10700 CPU without GPU. Before inference, the latent variables of hGP-LVM and sparse hGP-LVM and the variational mean of Bayesian hGP-LVM were initialized by generating two-dimensional random variables following $U(-10^{-3}, 10^{-3})$ and then mapping
them into the Lorentz model as $[\sqrt{1+x_{1}^{2}+x_{2}^{2}},x_{1},x_{2}]^{\top}$. We initialized the variance parameter $U(-10^{-5}, 10^{-5})$ in the Bayesian hGP-LVM. During optimization, we set a large learning rate at the beginning of optimization and fixed variance parameters in the first $100$ epochs for Bayesian hGP-LVM. The position of inducing variables $\bm{\mathrm{Z}}$ was updated every $10$ epoch by sampling from latent variables. Our implementation is available in \url{https://github.com/koshi-lmd/hyperboloid_gplvm.git}.
\par
We next notified the details of the quality metrics we used. Trustworthiness $T(k)\in[0,1]$ is a local one and defined as follows:
\begin{align}
    T(k)=1-\frac{2}{Nk(2N-3k-1)}\sum_{i=1}^{N}\sum_{j\in U_{i}^{k}}\mathrm{max}(0,r(i,j)-k)\label{eq_apx: trustworthiness},
\end{align}
where $U_{i}^{k}$ denotes a set of the $k$-nearest neighbors of the sample $i$ in the \textit{latent} space, and $j$ is an $r(i,j)$-th neighbor of $i$. Continuity $C(k)$ is the converse of trustworthiness computed by changing $U_{i}^{k}$ in Eq.~\eqref{eq_apx: trustworthiness} into $V_{i}^{k}$, a set of the $k$-nearest neighbors of the sample $i$ in the~\textit{observed} space. The Shepard goodness is the Spearman rank correlation in the Shepard diagram. The Shepard diagram is the scatterplot of the pointwise distance between two variables, and there are $\frac{N(N-1)}{2}$ points corresponding to the distance value between any two variables. The quality metrics for DR are compared in~\citep{espadoto2019toward}.
\section{Additional Experimental Results}
\begin{figure}[p]
    \centering
    \includegraphics[width=170mm]{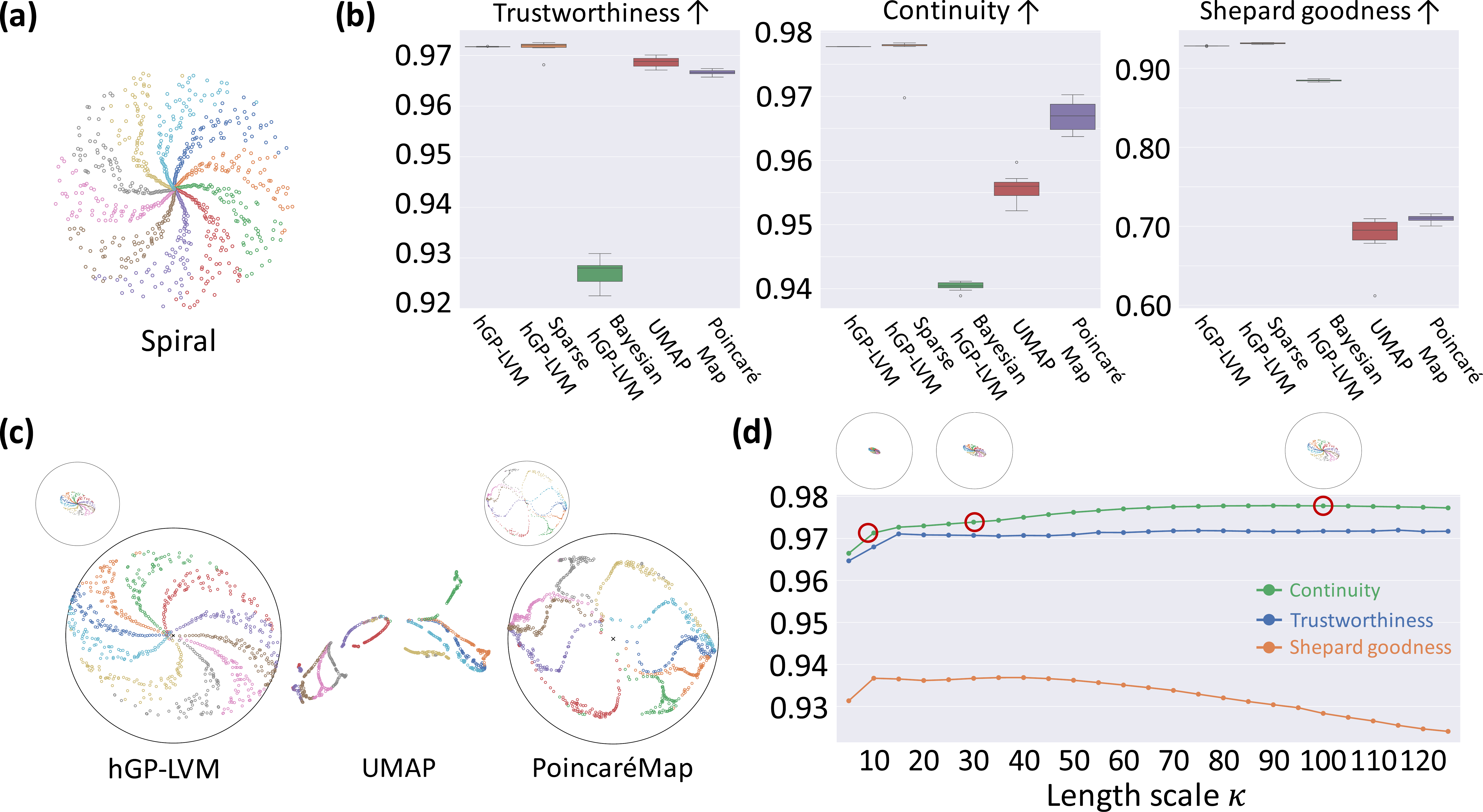}
    \caption{\textbf{Experimental results on the Spiral dataset.} (a) Shape of the spirals before linear transformation. (b) Error bar plot of trustworthiness (left), continuity (center), and Shepard goodness (right). (c) Scatter plot of two-dimensional embeddings of hGP-LVM, UMAP, and Poincar{\'e}Map. We zoomed in on latent variables and showed the original latent space in the upper left corner. (d) Quality metric scores of hGP-LVM with different length scales $\kappa$. (e) Comparison of reconstruction error of sparse hGP-LVM and sparse GP-LVM. }
    \label{fig:spiral_entire_results}
    \vfill
\end{figure}

In this section, we show the additional experimental results to compare our methods with the visualization-aided DR methods using synthesis datasets.
\paragraph{Spiral.}
First, we validate the effectiveness of the GP to preserve the continuity with the \textit{Spiral} dataset. The Spiral dataset contains $10$ spirals starting from the origin with random oscillation in proportion to the norm (Figure~\ref{fig:spiral_entire_results} (a)), synthesizing easy hierarchies. We generated $10$ spirals with $80$ points in a two-dimensional space ($N=800$) and mapped them in a $20$-dimensional space through a random linear transformation ($D=20$). We embedded the Spiral into the two-dimensional space and compared the quality with quantitative metrics and qualitative visualization. For quantitative evaluation, we used the three metrics following~\citep{espadoto2019toward, zu2022spacemap} to evaluate the preservation quality of the local and global structure: trustworthiness~\citep{venna2001neighborhood}, continuity, and Shepard goodness~\citep{joia2011local}. Trustworthiness tells the reliability of the embedding, and continuity measures the preservation quality of the local continuity. The local metrics require the number of neighbors, and we set $k=3$ in both metrics. The Shepard goodness is a global metric and the Spearman rank correlation of the pointwise distances between observed and latent variables. The Shepard goodness measures the match of global coordinates between variables. Since we expected the visualization purposes, we computed each metric using Euclidean coordinates. We compared hGP-LVMs with UMAP~\citep{mcinnes2018umap} as a benchmark for DR, and Poincar{\'e}Map~\citep{klimovskaia2020poincare} as a DR method on the hyperbolic model to confirm the effectiveness of the GP-based modeling. We set $\kappa=100$ and $M=30$ in all hGP-LVMs and conducted the same experiment ten times to confirm the reproducibility. 
\par

\begin{figure}[p]
    \centering
    \includegraphics[width=170mm]{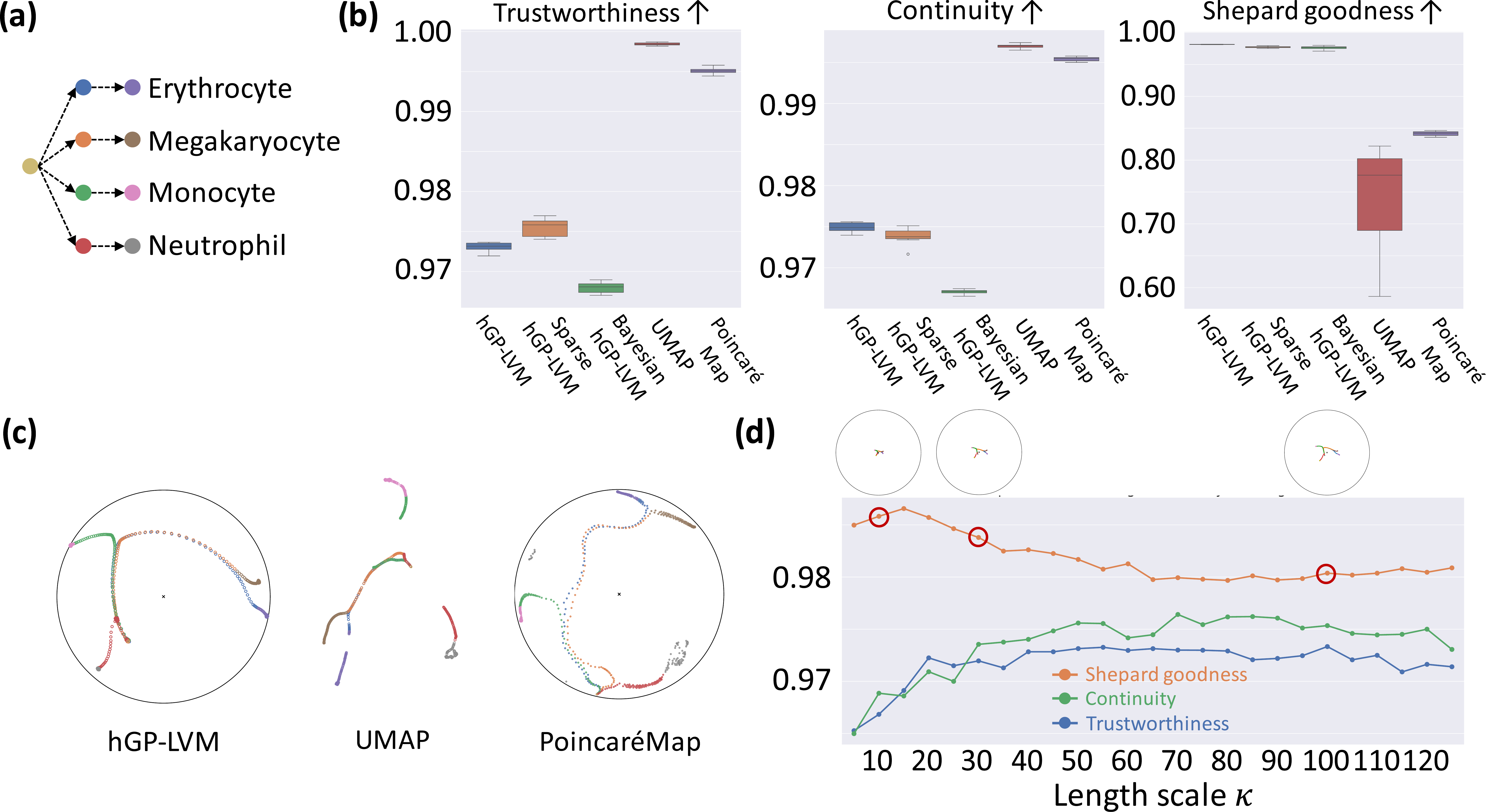}
    \caption{\textbf{Experimental results on the synthetic myeloid progenitors dataset.} (a) Hierarchical relation and color code of visualization. (b) Error bar plot of trustworthiness (left), continuity (center), and Shepard goodness (right). (c) Scatter plot of two-dimensional embeddings of hGP-LVM, UMAP, and Poincar{\'e}Map. We zoomed in on latent variables and showed the original latent space in the upper left corner. (d) Quality metric scores of hGP-LVM with different length scales $\kappa$. (e) Comparison of reconstruction error of sparse hGP-LVM and sparse GP-LVM.}
    \label{fig:syn_myeloid_entire_results}
\end{figure}

We first show the quantitative results in Figure~\ref{fig:spiral_entire_results} (b). hGP-LVM and sparse hGP-LVM performed the highest results with low variances in all metrics, indicating that they preserved Spirals' global and local structure. However, Bayesian hGP-LVM contains high variance in the local metrics caused by the approximated inference. Figure~\ref{fig:spiral_entire_results} (c) shows that UMAP and Poincar{\'e}Map preferred the neighbor relations of Spiral, unlike that of hGP-LVM, which preferred their global coordinates. The results in Figures~\ref{fig:spiral_entire_results} (b) and (c) imply that local and global structure preservation is needed to
embed the continuity of structured data. Figure~\ref{fig:spiral_entire_results} (d) shows the detailed results to confirm the effect of the length scale parameter $\kappa$. The range of the latent variables spread with large $\kappa$, and the continuity score increased, which comes in the hyperbolic curvature of the latent space. However, the Shepard goodness score decreased with large $\kappa$, and we observed the tradeoff relationship between global and local preservation quality. From the above, we confirm that hGP-LVMs have higher results than comparatives in the Spiral dataset.
\paragraph{Synthetic Myeloid Progenitors~\citep{klimovskaia2020poincare}.} 
We show another synthetic experimental results on the Myeloid Progenitors (MP) dataset, synthesizing cell differentiation from the progenitors into four cells: erythrocyte, neutrophil, monocyte, and megakaryocyte $(N=640,\,D=11)$ (Figure~\ref{fig:syn_myeloid_entire_results} (a)). We set $\kappa=100$ and $M=50$. 
\par
Figure~\ref{fig:syn_myeloid_entire_results} (b) shows the quantitative results. hGP-LVM outperformed the comparative methods except for the local metrics. However, the absolute value was still high, and hGP-LVM produced high accuracy in both local and global metrics. Figure~\ref{fig:syn_myeloid_entire_results} (c) shows the visualization results of hGP-LVM, GP-LVM, and Poincar\'{e}Map. Although the embeddings of GP-LVM and Poincar\'{e}Map are torn or wiggling, the hGP-LVM well preserved the continuity behind the hierarchical data, contributing to the visibility of the low-dimensional embeddings. The effect of the length scale in Figures~\ref{fig:spiral_entire_results} (d) is similar to the result on the Spiral dataset in Figure~\ref{fig:syn_myeloid_entire_results} (d). From the above, we confirm that hGP-LVMs produced better results than previous visualization-aided DR methods in the synthetic setting.

\subsection{Embedding Comparison with Different Lengthscale}
\label{appendix:different_lengthscale}

\begin{figure}[p]
    \centering
    \includegraphics[width=110mm]{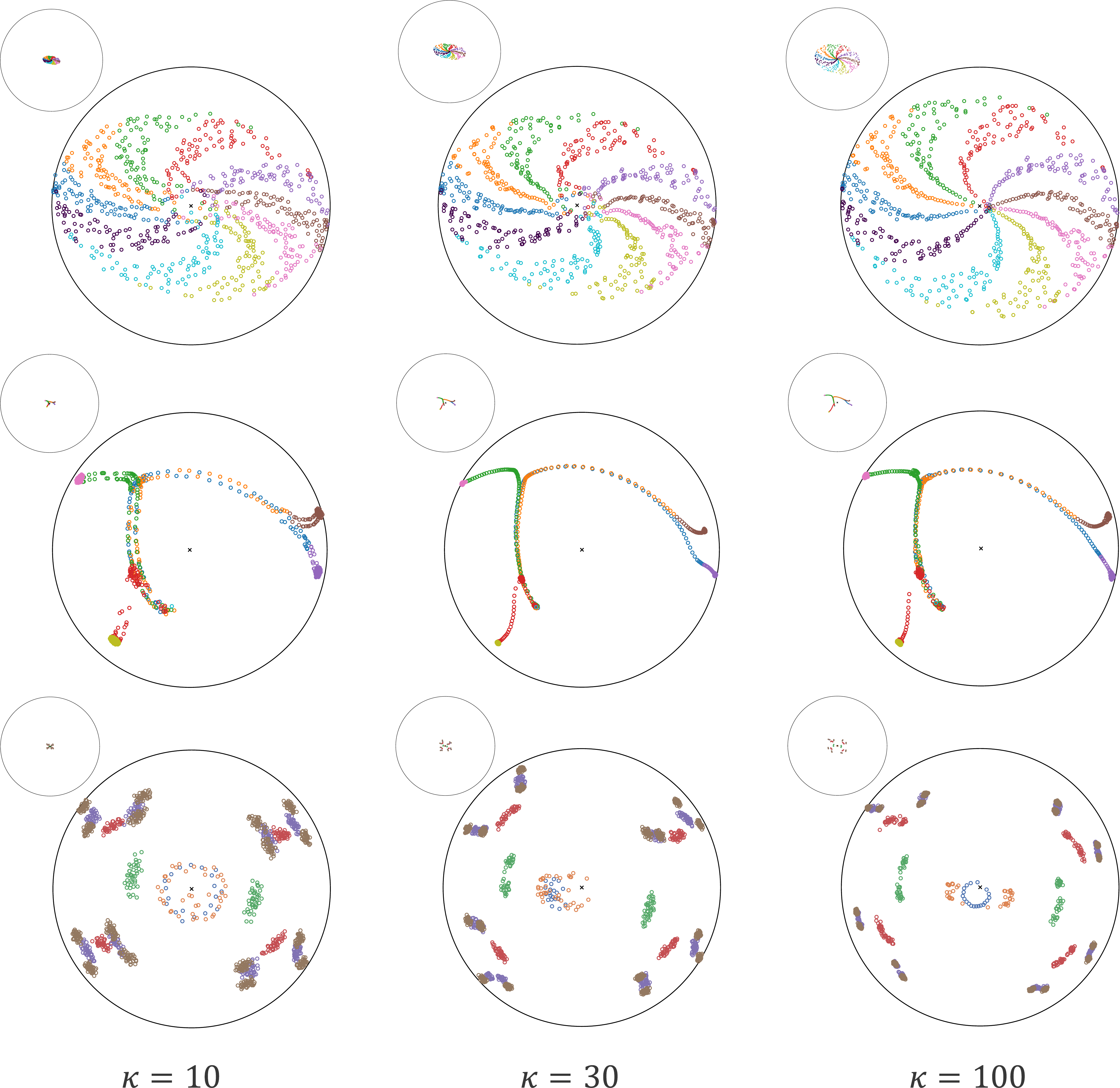}
    \caption{\textbf{Embedding comparison of hGP-LVM with different length scales on synthetic datasets, Spiral (top), synthetic myeloid progenitors (middle), and SBT (bottom)).} We zoomed in on the latent variables.}
    \label{fig:lengthscale}
\end{figure}
This section presents the visualization comparison with different length scales, $\kappa$. In Section 3.1 of the main paper, we state that \textit{the meaning of the length scale $\kappa$ is how much we expect the latent variables to follow the hyperbolic curvature}. We show the experimental verification of this statement in Figure~\ref{fig:lengthscale} on all synthesis datasets. In all results, we can confirm that the representation was spread and aligned as growing $\kappa$. In the SBT results (bottom), although the embeddings of depths $1$ and $2$ were mixed around the origin, they were separated with $\kappa=100$. The length scale parameter determined the range of the latent variables, and large $\kappa$ brought the hyperbolic curvature to the latent representation. We must determine $\kappa$ according to the degree to which we expected the hierarchical structure in data.

\end{document}